\pdfoutput=1

\documentclass[11pt]{article}

\usepackage{acl2023}

\usepackage[utf8]{inputenc} 
\usepackage[T1]{fontenc}    
\usepackage{hyperref}       
\usepackage{url}            
\usepackage{booktabs}       
\usepackage{amsfonts}       
\usepackage{nicefrac}       
\usepackage{microtype}      
\usepackage{xcolor}         
\usepackage{xspace}

\usepackage{listings}

\usepackage{array}

\usepackage{tcolorbox}
\usepackage{multicol}
\usepackage{times}
\usepackage{latexsym}
\usepackage{import}
\usepackage{graphicx}
\usepackage{booktabs}
\usepackage{amssymb}
\usepackage{multirow}
\usepackage{blindtext}
\usepackage{enumitem}
\usepackage{xcolor}
\usepackage{xparse}
\usepackage{longtable}
\DeclareDocumentCommand\dia{ o m }{%
    \begin{itemize}[%
        ,label=\IfNoValueTF {#1} {}{#1:}
        ,labelsep=8mm
        ,nosep 
        ,font=\color{black}
        ]
        \item #2
    \end{itemize}%
    }

\definecolor{darkgreen}{rgb}{112,114,88}
\definecolor{goldanswer}{HTML}{0072B2}
\definecolor{deraanswer}{HTML}{009E73}
\definecolor{oneshotanswer}{HTML}{D55E00}

\usepackage[T1]{fontenc}

\usepackage[utf8]{inputenc}

\usepackage{microtype}
\newcolumntype{P}[1]{>{\centering\arraybackslash}p{#1}} 

\usepackage{inconsolata}

\newcommand*{\LLM}{GPT-4}

\newcommand{\eg}{\textit{e.g,}\xspace}
\usepackage{tcolorbox}
\tcbset{
  width=1\textwidth,
  halign=justify,
  center,
  colback=white    
}

\usepackage{xcolor}
\usepackage{outlines}

\definecolor{codegreen}{rgb}{0,0.6,0}
\definecolor{codegray}{rgb}{0.5,0.5,0.5}
\definecolor{codepurple}{rgb}{0.58,0,0.82}
\definecolor{backcolour}{rgb}{0.95,0.95,0.92}

\lstdefinelanguage{promptlanguage}{
    morecomment=[l][\color{codepurple}]{//},
    morecomment=[s][\color{blue}]{\{}{\}},
}
\newcolumntype{M}[1]{>{\centering\arraybackslash}m{#1}}

\lstdefinestyle{promptstyle}{
    backgroundcolor=\color{white},   
    commentstyle=\color{codegreen},
    keywordstyle=\color{magenta},
    numberstyle=\tiny\color{codegray},
    stringstyle=\color{codepurple},
    basicstyle=\ttfamily\small,
    frame = single,
    breakatwhitespace=false,         
    breaklines=true,                 
    captionpos=b,                    
    keepspaces=true,                 
    numbers=left,                    
    numbersep=5pt,                  
    showspaces=false,                
    showstringspaces=false,
    showtabs=false,                  
    tabsize=2
}

\newenvironment{chat} {
    
    \newcommand\who[1]{\par\hangindent=2.5em\hangafter=1 ##1:}
    \par\vskip2em
}{%
    \par
    \hangindent=0pt\hangafter=0
    \vskip2em
}

\newcommand{\methodname}{DERA}
\newcommand{\researcher}{\textit{Researcher}\xspace}
\newcommand{\decider}{\textit{Decider}\xspace}

\newcommand{\Researcher}{\textit{Researcher}\xspace}
\newcommand{\Decider}{\textit{Decider}\xspace}

\title{DERA: Enhancing Large Language Model Completions \\ with  Dialog-Enabled Resolving Agents}

\author{Varun Nair\thanks{~~The first two authors contributed equally to this work.} \hspace{1em}
        Elliot Schumacher$^*$ \hspace{1em}
        Geoffrey Tso \hspace{1em}
        Anitha Kannan \hspace{1em}\\ 
        \texttt{\{varun, elliot, geoff, anitha\}@curai.com} \\
        Curai Health 
        }

\begin{document}

\maketitle

\begin{abstract}
Large language models (LLMs) have emerged as valuable tools for many natural language understanding tasks. In safety-critical applications such as healthcare, the utility of these models is governed by their ability to generate outputs that are  factually accurate and complete. In this work, we present dialog-enabled resolving agents (DERA). DERA is a paradigm made possible by the increased conversational abilities of LLMs, namely GPT-4.  It provides a simple, interpretable forum for models to communicate feedback and iteratively improve output. We frame our dialog as a discussion between two agent types -- a \Researcher, who processes information and identifies crucial problem components, and a \Decider, who has the autonomy to integrate the \Researcher's information and makes judgments on the final output.

We test \methodname~against three clinically-focused tasks. For medical conversation summarization and care plan generation, DERA shows significant improvement over the base GPT-4 performance in both human expert preference evaluations and quantitative metrics. In a new finding, we also show that GPT-4's performance (70\%) on an \textit{open-ended} version of the MedQA question-answering (QA) dataset (\citet{jin2021disease}, USMLE) is well above the passing level (60\%), with \methodname~showing similar performance. We release the open-ended MEDQA dataset at \url{https://github.com/curai/curai-research/tree/main/DERA}.
\end{abstract}

\begin{figure*}[ht!]
    \centering
    \includegraphics[width=1.0\textwidth]{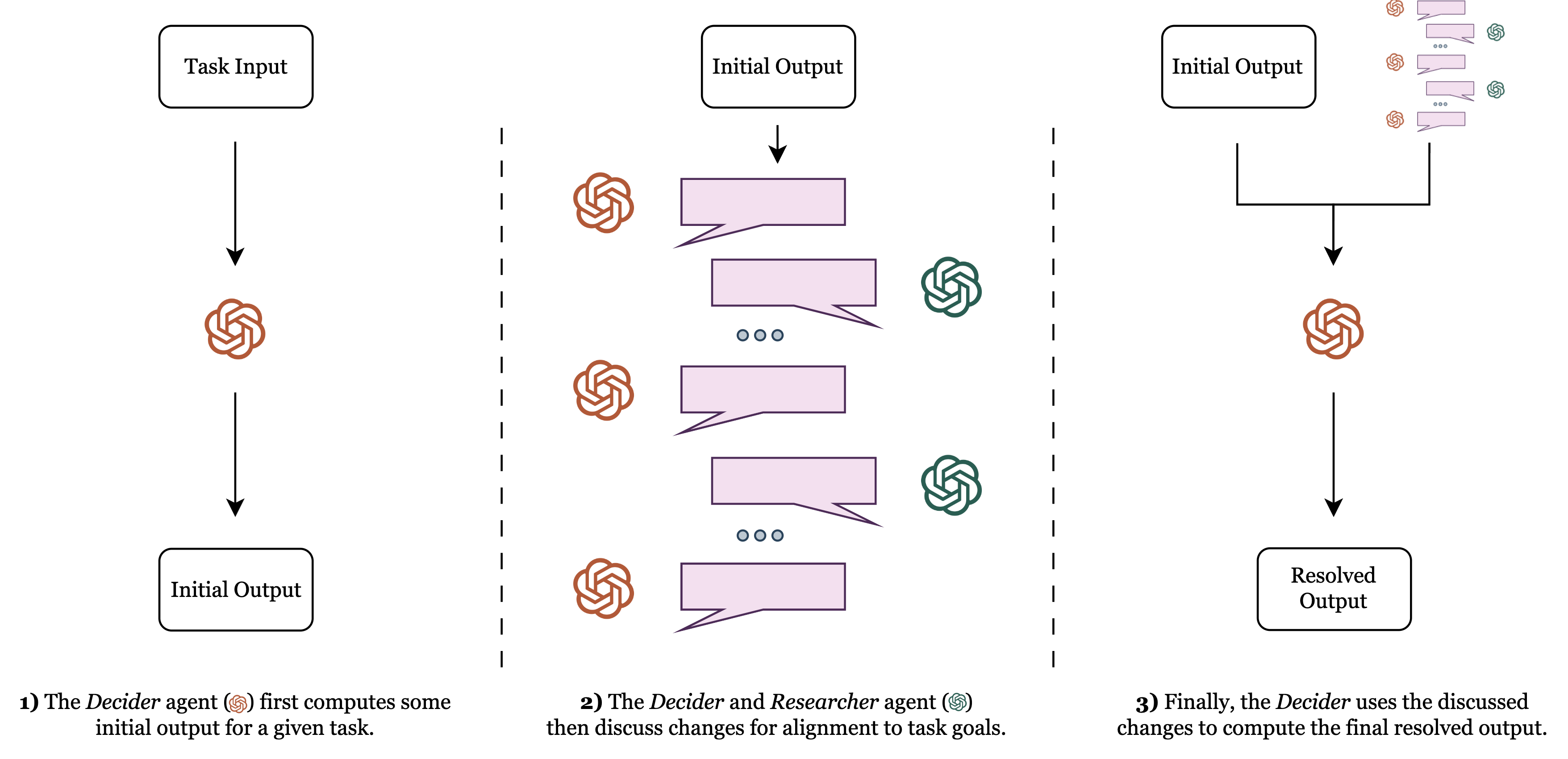}
    \caption{Overview of DERA. The method consists of two agents--a \Researcher and a \Decider. The \Decider generates an initial output for the task (step 1).  Then, the \Decider and \Researcher work through the problem via conversation (step 2), with the \Researcher tasked to help identify crucial problem components. The \Decider has the autonomy to integrate the \Researcher’s inputs and makes judgments on the final output (step 3). Neither agent has knowledge of the ideal final output. 
    }
    \label{fig:dera}
\end{figure*}

\section{Introduction}
Large language models (LLMs; \citet{brown2020gpt3, lewis-etal-2020-bart}) are deep-learning models that have been trained to predict natural language text conditioned on an input.  The use of these models has led to advances in natural language performance far beyond just language modeling tasks.  Within the realm of medicine, LLM-powered methods have shown improvements in medical tasks such as question answering \cite{singhal2022large,lievin2022can}, information extraction \cite{Agrawal2022LargeLM}, and summarization \cite{chintagunta-etal-2021-medically}.  

LLM-powered methods use natural language instructions called \textit{prompts}.  These instruction sets often include a task definition, rules the predictions must follow, and optionally some examples of the task input and output \cite{reynolds2021prompt, brown2020gpt3}. The ability of generative language models to create output based on natural language instructions (or prompts) removes the need for task-specific training \cite{min-etal-2020-demonstrations} and allows non-experts to build upon this technology.  

While many tasks can be formulated as a single prompt, later work has shown that breaking down single tasks into sub-tasks (called \textit{chaining}) has benefits in terms of task performance and interpretability \cite{wu2022AIChains}. Examples of chaining strategies include chain-of-thought \cite{wei2022chain} and other task-specific approaches (\eg \citet{Agrawal2022LargeLM}).  Chain-of-thought strategies  prompt the model to think through a problem as an expert might approach it, leading to improvements in some tasks \cite{lievin2022can,wang-etal-2022-iteratively,tafjord-etal-2022-entailer}. 

All of these approaches attempt to coerce the correct generation from the base language model.  However, one fundamental limitation of this strategy is that these prompting architectures are restricted to a fixed set of prompts designed for specific tasks in mind, such as writing explanations or resolving anomalies within the output.  Furthermore, they struggle with generating factually accurate text and often can include hallucinations and omissions \cite{maynez-etal-2020-faithfulness,dziri-etal-2022-origin,berezin-batura-2022-named}.  This poses a significant hurdle when applying them to real-world scenarios, especially in the clinical domain.

We advocate for a different approach that has two essential elements.  First, it consists of an iterative approach to refining the initial output. This allows the generation to be refined holistically as opposed to conditional chaining.  Second, it includes an advisor that can guide by suggesting areas to focus on in each iteration, adding interpretability to the process. With the advent of \LLM~\cite{openai2023gpt4} capable of robust, realistic conversation, we can use dialog as the medium for interaction. 

We propose DERA: Dialog-Enabled Resolving Agents. DERA is a framework to explore how we can improve performance on natural language tasks using agents tasked with resolving (or improving) the output through dialog.  We propose that scoping each agent in the dialog to a specific role will better enable them to focus on discrete portions of the task, and ensure their partner agent stays aligned with the overall goal.  One agent role, the \Researcher, works to identify relevant information for the problem and suggest areas of focus to the other agent.  Another agent role, the \Decider, has the autonomy to react to that information and make final decisions about the output.

Our paper makes the following contributions:
\begin{outline}[itemize]
    
    \1 We introduce DERA (\S~\ref{sec:dera}) - a framework for agent-agent dialog to improve performance on natural language tasks.
    \1  We evaluate DERA on three different types of clinical tasks.  Each of these requires different types of textual inputs and types of knowledge to solve. 
    \2 The medical conversation summarization task (\S~\ref{sec:summarization}) focuses on generating a summary of a doctor-patient chat that is factually accurate without omissions or hallucinations.  
    \2 The careplan generation task (\S\ref{sec:careplan})  is knowledge-intensive with long outputs that are useful in clinical decision support. There is not a single best answer to the task and the goal is to maximize the amount of factually accurate and relevant information generated.
    \2 Medical question answering \cite{jin2021disease} is a knowledge reasoning task with a single answer but posed as an open-ended task without access to multiple choices. We study in this harder setting using two question-answering datasets (\S\ref{sec:qa}). 
    \1 In both human-annotated evaluations, we find that \methodname~outperforms base \LLM~ performance in the careplan generation and medical conversation summarization tasks on a variety of metrics. In quantitative evaluations, we find that DERA successfully corrects medical conversation summaries with large amounts of errors. Conversely, we find small to no improvement between \LLM~performance and \methodname~on question-answering.
    \1 We theorize this approach is well suited for longer-form generation tasks, in which there are a lot of fine-grained details.
    \1 We will work to release a new open-ended medical question-answering task based on MedQA, which consists of United States Medical Licensing Exam (USMLE) practice questions\footnote{\url{https://github.com/curai/curai-research/tree/main/DERA}}. This opens up new research in the modeling and evaluation of question-answering systems.
\end{outline}

\section{\methodname: Overview}
\label{sec:dera}
\methodname~ is a general chat framework that leverages dialog-capable agents to iteratively work through a task (Figure \ref{fig:dera}). We focus on agent setups that work to probe knowledge sources, whether internal (from within \LLM) or external (from text, documents, etc.).  In approaches like chain-of-thought, these roles are generally performed jointly.  In contrast, we propose that pairing an information-focused agent with a decision-maker agent will lead to a higher-quality output.  Furthermore, this approach allows for \methodname~ to alternate between processing knowledge and acting upon them, as opposed to doing them concurrently.  

First, we propose the use of a \Researcher agent. The goal of a researcher agent is to review pieces of information -- which can be internal to \LLM~or external -- and make suggestions on what is likely to be crucial in solving the problem.  As we do not have a definitive source of what is and is not relevant, we rely on \LLM 's ability to identify relevancy in light of the current task.  We do not treat this agent as the definitive source of truth.  Rather, we task it with being helpful and constructive during the dialog.

Second, we propose the use of a \Decider agent. In addition to starting the conversation, this agent is tasked with responding to the information provided by the \Researcher agent, and deciding whether to integrate that information into the task output.  This allows \LLM~to make discrete decisions in reaction to the information highlighted by the \researcher.  At no point, however, does the \decider defer to the \researcher. This agent is ultimately responsible for the final decision, and while it is tasked with reviewing all information highlighted by \researcher, it does not have to use any of that information.

The specifics of each agent can vary for different tasks. For Question Answering, the \researcher is tasked with pulling information from the question, using the internal knowledge of \LLM~alone.  For summarization, the \researcher has access to external texts which contain the full patient encounter.  Conversely, the edits to the text generation task are made incrementally by the \textit{Decider} in the summarization task, while they are made more discretely in the question-answering task.  In some settings, agents take a hybrid role, where they each have access to different information and jointly make decisions.  Overall, the  goal remains the same -- that this approach allows for information to be processed in a manner that is both role-defined and iterative, producing better quality output. 

We apply \methodname~to three natural language generation tasks.  The first, medical conversation summarization (\S\ref{sec:summarization}), probes the ability of \methodname~ to create a summary of a doctor-patient chat.  This requires the ability to identify and rewrite medically-relevant information in a concise format.  The second, care plan generation (\S\ref{sec:careplan}), tests whether \methodname~can generate doctor-facing suggestions for potential actions to address patient concerns.  This requires similar abilities, with the added challenge of knowing the appropriate next steps for a variety of medical conditions.  Finally, medical question-answering (\S\ref{sec:qa}) tests the ability of \methodname~to generate a wide variety of medical knowledge in a short format.

\section{Medical Conversation Summarization}
\label{sec:summarization}
\paragraph{Overview} The task of medical conversation summarization is to encapsulate a patient-doctor conversation \citep{enarvi-etal-2020-generating, joshi-etal-2020-dr, zhang2021leveraging, chintagunta-etal-2021-medically}. We focus on summarizing patient-doctor chats into six independent sections: \textit{Demographics and Social Determinants of Health}, \textit{Medical Intent}, \textit{Pertinent Positives}, \textit{Pertinent Negatives}, \textit{Pertinent Unknowns}, and \textit{Medical History}.   This structured format requires the model to summarize the chat while placing each piece of information in the appropriate section.   As these summaries are typically used by doctors for downstream tasks such as clinical decision-making, it is important that the generated summaries are both factually accurate (no hallucinations) and complete (no omissions).

\paragraph{\methodname~Setup} 
We formulate the \methodname~setup for medical conversation summarization as follows.
Both  \Decider  and \Researcher have access to the full medical conversation  between the patient and the physician. Both agents are prompted to converse with one another. The \Decider agent generates an initial summary of the medical conversation (Prompt \ref{prompt:summarization}) and shares it with the \Researcher agent. The \Researcher agent's role (Prompt \ref{prompt:summ_researcher})  is to ``read'' the summary and  point out any discrepancies to \Decider.  \Decider, using  Prompt \ref{prompt:summ_decider}, either accepts or rejects those discrepancies, by agreeing with the suggestion or disagreeing and responding with some reasoning. Instead of regenerating the summary at each step of the conversation, \Decider writes the accepted suggestions to a shared \textit{scratchpad}, which acts like a memory that it uses at the end of the conversation to generate the final summary. The conversation terminates once \Researcher is satisfied with the suggestions made to the scratchpad or a maximum conversation length is reached (set to 15 turns total). As the final step, the \Decider generates (Prompt \ref{prompt:summ_final}) the final summary using the contents of the scratchpad and the original summary.

 GPT-4 prompts are run with the settings mentioned in Table \ref{tab:summ_experimental_details}. 

\begin{figure*}[ht!]
    \centering
    \includegraphics[width=1.0\textwidth]{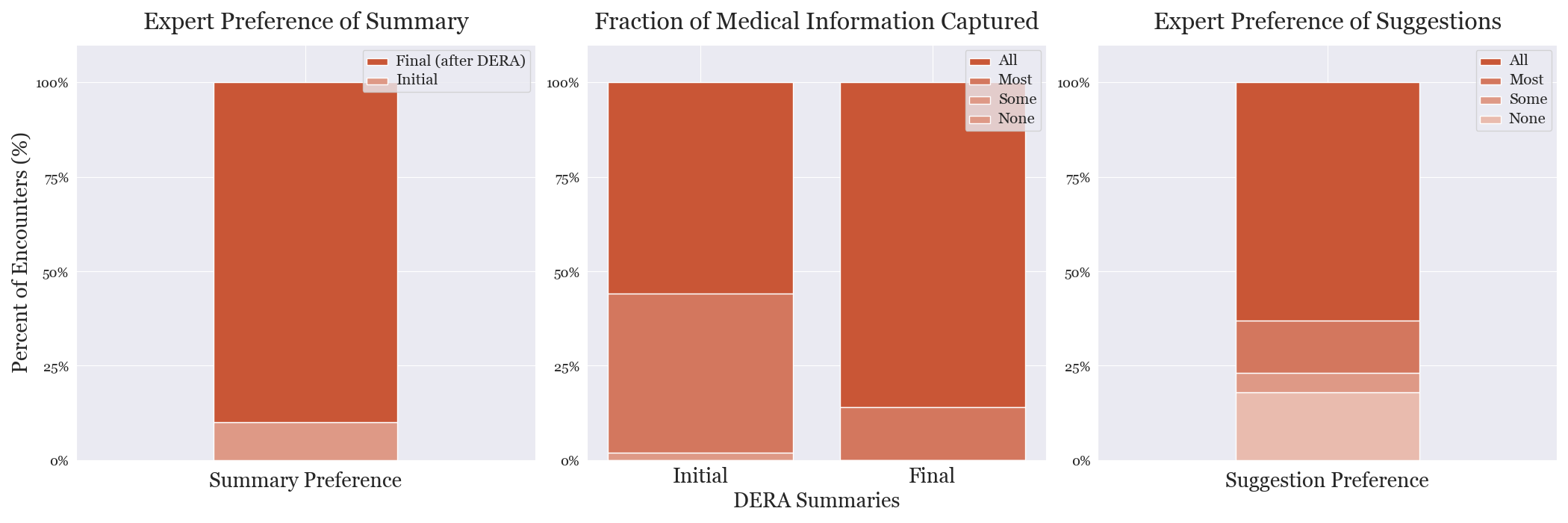}
    \caption{Results from physician-expert evaluations on the medical conversation summarization task. (Left) Physicians choose the final summary produced by DERA over the initial GPT-4 generated summary 90\% to 10\%. (Center) Final DERA summaries capture far more clinical information than initial GPT-4 generated summaries, with physicians rating "All" relevant clinical information from the patient-physician chat captured in 86\% of DERA summaries vs. 56\% of initial GPT-4 summaries. (Right) For summary correction suggestions in the scratchpad, physicians rate agreement with "All" suggestions in 63\% of encounters, Most" in 14\%, "Some" in 5\%, and "None" in 18\%.}
    \label{fig:results_summ}
\end{figure*}
\begin{table*}[ht!]
    \centering
    \resizebox{1\textwidth}{!}
    {
\begin{tabular}{P{1.5cm}P{1.5cm}P{1.8cm}P{1.8cm}P{1.8cm}P{1.5cm}M{1.5cm}}
\toprule

Corruption Level & Summ. Version & Pertinent Positives & Pertinent Negatives & Pertinent Unknowns & Medical History & Average\\ \hline
\multirow{2}{*}{low ($\frac{3}{10}$)} & {\small Initial} & 89.38 & 83.05 & 87.42 & 80.88 & 85.18 \\
 & {\small DERA} & 95.65 & 96.77 & 97.10 & 97.35 & \textbf{96.71} \\ \bottomrule
\multirow{2}{*}{medium ($\frac{5}{10}$)} & {\small Initial} & 83.12 & 81.6 & 71.14 & 73.82 & 77.42 \\
 & {\small DERA} & 94.29 & 95.31 & 96.17 & 98.12 & \textbf{95.97} \\ \bottomrule
\multirow{2}{*}{high ($\frac{7}{10}$)} & {\small Initial} & 68.35 & 70.07 & 68.79 & 57.27 & 66.12 \\
 & {\small DERA} & 92.96 & 90.86 & 94.81 & 95.16 & \textbf{93.45} \\ \bottomrule
\end{tabular}
}
    \caption{Medical conversation summarization task: Quantitative evaluation (GPT-F1 scores) of the initial  summary with errors and the \methodname~corrected version. We show that by introducing synthetic corruption (hallucinations, omissions, etc.) into medical summaries, DERA can resolve these corruptions at low, medium, and high levels of corruption. GPT-F1 scores for the DERA-produced summary are consistently higher than the initial summaries.}
    \label{table:corruption}
\end{table*}

\paragraph{Dataset} 
We randomly sampled 500 medical encounters from a chat-based telehealth platform. Each encounter contains the patient's age, sex, and chat conversation with a licensed medical provider. Encounters in this dataset cover a wide variety of common presentations in telehealth, including urinary tract infections, back/abdominal pains, toothaches, and others. All data is de-identified and scrubbed for protected health information prior to experimentation. Conversations contain 27 dialog turns on average (min of 9 turns, max of 82 turns) and average 646 unigram tokens per encounter (min 42 tokens, max 2031 tokens).

\paragraph{Human Expert Evaluation}

To evaluate the effectiveness of DERA to generate better summaries, we conducted human evaluation studies with four licensed physicians on a random subset of 50 out of the 500 encounters described above. We sampled a smaller, random subset due to the high labeling cost induced by using expert physicians.  

The licensed physicians were provided with the encounter and the two summaries.  These included the initial GPT-4 generated summary, and the final generated summary produced using DERA. Each physician was asked to answer three main questions in the light of the summary's clinical utility for themselves or another physician:
\begin{enumerate}
    \item Which summary do you prefer to use for the given patient and encounter?  (Options: Initial, Final).
    \item What percentage of the overall clinical information in the dialog is captured by the summary? (Options: All, Most, Some, None)
    \item What percentage of the suggestions added to the DERA scratchpad do you agree with? (Options: All, Most, Some, None)
\end{enumerate}

Figure \ref{fig:results_summ} shows the results of our human expert evaluation. Physicians notably choose the summary produced after DERA over the initially generated summary 90\% - 10\%.  Their preference for the DERA-produced summary is further corroborated by the fraction of medical information captured in the final DERA summary vs. initial, as final summaries were rated as capturing "All" medical information from the patient-physician dialog in 86\% of encounters vs. the initial summaries  capturing "All" medical information in just 56\% of encounters.   In general, we also find broad agreement for the suggestions in each encounter's scratchpad: they agreed with "All" corrections suggested for a given encounter's summary 63\% of the time, "Most" 14\% of the time, "Some" 5\% of the time, and "None" 18\% of the time. On average, each scratchpad contains 2-3 suggestions.

In addition to these questions, we also asked the physician-experts the following: \textit{If this summary were acted upon by another clinical provider, does this summary contain information that could potentially be harmful to the patient given their presentation?} (Options: Yes, No). The amount of summaries containing "harmful" information drops from 2\% in the initial summary to 0\% in the final DERA summary.  We caution against drawing generalizations from these harmfulness numbers. Our evaluations are both limited in number and drawn from a patient population specific to the telehealth platform; thus cannot predict the generalizability of these findings in other settings.

\paragraph{Quantitative Evaluation}
\label{sec:summ_quan_eval}
We also perform a more large-scale study without the need for human annotation. We generate GPT-4 summaries for all the 500 encounters and assume them to be ground truth. Then, we synthetically induce ``corruptions'' into the generated summary and use that as the initial input. These mistakes artificially lower the summary's quality and produce significant hallucinations and omissions.  The goal is to quantitatively evaluate DERA's ability to write medical summaries by measuring the degree to which the \Researcher and \Decider agents can identify and fix "corruptions" introduced to the medical summary. 

Prompt \ref{prompt:summ_corruption} contains specific instructions for generating the corruptions. We can control the level of corruption desired by passing one of three levels of corruption as a variable to our corruption prompt: low ($\frac{3}{10}$), medium ($\frac{5}{10}$), or high ($\frac{7}{10}$). The higher the corruption, the more symptoms could be rearranged.  Similarly, hallucinated symptoms could be introduced, among other corruptions. For a qualitative example of this process of generating an initial summary, corrupting it, resolving with DERA, and generating a final summary see Fig.~\ref{fig:summ_ex}.

We measure the degree to which corruptions are present by using a GPT-based metric that tracks the medical concept coverage of the medical summary, \textbf{GPT-F1}. To compute GPT-F1, we compute the harmonic mean of two sub-metrics: GPT-Recall and GPT-Precision. We describe each sub-metric below.

\textbf{GPT-Recall}: To compute, we first extract medical entities from both the predicted text and ground-truth text\footnote{Note that the terms "predicted" and "ground-truth" are used loosely here, and can also refer to synthetic text that is used as ground-truth (as in the case of the corruption experiments).} of the same summary section (using Prompt \ref{prompt:metric_extractor}) and use a verification prompt (Prompt \ref{prompt:metric_verifier}) to infer if the entities extracted from the ground-truth section are also present in the predicted text, This produces $tp_{gt}$ and $f_{n}$ values, which is used to calculate $\textnormal{GPT-Recall} = \frac{tp_{gt}}{tp_{gt}+f_{n}}$.

\textbf{GPT-Precision}: To compute, we also first extract medical entities from the corresponding predicted and ground-truth summary sections and verify concepts extracted from the predicted section are also present in the ground-truth text, either as exact matches or re-phrasings. This produces $tp_{pred}$ and $f_{p}$, which is used to calculate $\textnormal{GPT-Precision} = \frac{tp_{pred}}{tp_{pred} + f_{p}}$.

We present the results of our quantitative evaluation using the GPT-F1 metric in Table \ref{table:corruption}. Specifically, we compare GPT-F1 on the initial summary with errors to the \methodname~corrected summary.  Note first how the higher levels of corruption manifest in the initial summary GPT-F1. As the corruption level of the initial summary increases, the initial GPT-F1 score drops. We find that DERA can produce significantly improved summaries in low, medium, and high levels of corruption, as evidenced by increases in GPT-F1. This suggests that the interaction between the \Researcher and \Decider agents is identifying hallucinations and omissions and resolving them through dialog, even when many such corruptions are present.

\begin{figure*}[p]
\begin{tcolorbox}[boxsep=1pt,left=2pt,right=2pt,top=0pt,bottom=0pt]
\textbf{Doctor-Patient Chat}
\small
\vspace{-1em}

\begin{multicols}{2}
\begin{chat}
\who{Patient} UTI 
\who{Doctor} Hi NAME, thank you for starting a visit. My name is NAME, and I am your clinical associate for this live chat. \ldots I am sorry to hear that. 
\who{Patient} Okay thank you 
\ldots
\who{Doctor} When was your last sexual encounter? 
\who{Patient} Two days ago. I’m not having any since then because of the pain with my bladder infection.
\who{Doctor} Was the encounter unprotected? 
\who{Patient} No
\who{Doctor} How many sexual partners have you had in the past 6 months? 
\columnbreak
\who{Patient} Several. I did however get tested for HIV and STIs a couple weeks ago and all came back negative. I get tested regularly since I am sexually active.
\who{Doctor} Do you have a new sexual partner? 
\who{Patient} No new ones 
\who{Doctor} Do you douche or use feminine hygiene products? 
\who{Patient} No Sometimes I use baby wipes \\
\ldots
\who{Doctor} For what medical conditions you are taking BuSpar, LaMICtal, Zoloft? 
\who{Patient} Buspar for Bipolar, Lamictal for anxiety, Zoloft for depression 
\end{chat}
\end{multicols}
\vspace{-1em}
\end{tcolorbox}
\begin{tcolorbox}[boxsep=1pt,left=2pt,right=2pt,top=0pt,bottom=0pt]
\textbf{Initial Care Plan}
\vspace{0.25em}
\small

\textit{Tests}: (1) Perform a urinalysis and urine culture to confirm the presence of a urinary tract infection and to identify \ldots

\textit{Medications}: (1) Prescribe an appropriate antibiotic, such as nitrofurantoin, fosfomycin, or trimethoprim/sulfamethoxazole, to treat the infection. Note that the choice of antibiotic may be adjusted based on the results of the urine culture. (2) Continue the use of over-the-counter pain relief medication, such as AZO, until the antibiotic treatment relieves symptoms.

\textit{Lifestyle}: (1) Encourage the patient to increase fluid intake, particularly water, to help flush out bacteria from the urinary tract. (2) Suggest urinating frequently and fully emptying the bladder to help clear the infection. (3) Recommend proper hygiene practices, such as \ldots (4) Advise the use of a urinary health supplement \ldots

\textit{Referrals}: (1) If symptoms worsen or do not improve after a few days of antibiotic treatment, refer the patient to \ldots

\textit{Supportive Care}: (1) Provide education on recognizing the signs and symptoms of recurrent urinary tract infections and the importance of seeking timely medical care. (2) Offer reassurance and support regarding the patient's mental health and \ldots
\end{tcolorbox}

\begin{tcolorbox}[boxsep=1pt,left=2pt,right=2pt,top=0pt,bottom=0pt]
\textbf{\methodname~Dialog}
\small
\vspace{-1.75em}
\begin{chat}
\who{\Researcher} Starting with the Tests section, everything looks good with the urinalysis and urine culture being recommended. However, in the Medications section, it would be helpful to include that antibiotic choice should consider the patient's previous antibiotic exposure and allergies. It will also be useful to mention any potential drug interactions with BuSpar, LaMICtal, and Zoloft. In the Lifestyle section, there is no mention of the patient's sexual practices, which could contribute to recurrent UTIs. Considering the patient's history of multiple sexual partners, you might want to add a recommendation to educate the patient on safe sexual practices and the importance of using a barrier method during sex to reduce the risk of UTIs. \ldots

\who{\Decider} Thank you for your suggestions. I agree with the recommendation to consider the patient's previous antibiotic exposure and allergies, as well as potential drug interactions with their other medications. I also see the importance of addressing safe sexual practices and barrier method usage to reduce the risk of UTIs. In terms of supportive care, providing guidance on managing UTIs and potential medication interactions concerning their mental health medications is crucial.

\who{\Researcher} Thank you for considering my suggestions. Please make sure to discuss these points with the patient, and \ldots
\end{chat}
\vspace{-2em}
\end{tcolorbox}

\begin{tcolorbox}[boxsep=1pt,left=2pt,right=2pt,top=0pt,bottom=0pt]
\textbf{Scratchpad}
\small

1) Add a note in the Medications section to consider the patient's previous antibiotic exposure, allergies, and potential drug interactions with BuSpar, LaMICtal, and Zoloft. 

2) In the Lifestyle section, include a recommendation to educate the patient on safe sexual practices and the importance of using barrier methods during sex to reduce the risk of UTIs. 

3) In the Supportive Care section, offer guidance on managing UTIs and potential medication interactions related to their mental health medications.
\end{tcolorbox}
\begin{tcolorbox}[boxsep=1pt,left=2pt,right=2pt,top=0pt,bottom=0pt]
    \textbf{Final Care Plan}
\vspace{0.25em}
\small

\textit{Tests}: (1) Perform a urinalysis and urine culture to confirm the presence of a urinary tract infection and to identify \ldots

\textit{Medications}: (1) Prescribe an appropriate antibiotic, such as nitrofurantoin, fosfomycin, or trimethoprim/sulfamethoxazole, to treat the infection. Note that the choice of antibiotic may be adjusted based on the results of the urine culture. \textbf{Consider the patient's previous antibiotic exposure, allergies, and potential drug interactions with BuSpar, LaMICtal, and Zoloft.} (2) Continue the use of over-the-counter pain relief medication, such as AZO, until the antibiotic treatment relieves symptoms.

\textit{Lifestyle}: (1) Encourage the patient to increase fluid intake, particularly water, to help flush out bacteria from the urinary tract. (2) Suggest urinating frequently and fully emptying the bladder to help clear the infection. (3) Recommend proper hygiene practices, such as \ldots (4) Advise the use of a urinary health supplement \ldots \textbf{(5) Educate the patient on safe sexual practices and the importance of using barrier methods during sex to reduce the risk of UTIs.}

\textit{Referrals}: (1) If symptoms worsen or do not improve after a few days of antibiotic treatment, refer the patient to \ldots

\textit{Supportive Care}: (1) Provide education on recognizing the signs and symptoms of recurrent urinary tract infections and the importance of seeking timely medical care. (2) Offer reassurance and support regarding the patient's mental health \ldots \textbf{(3) Offer guidance on managing UTIs and potential medication interactions related to their mental health medications.}
\end{tcolorbox}
\caption{Qualitative example of care plan generation with DERA abridged for space. Starting with the initial chat, the \Decider generates an initial care plan. The \Researcher and \Decider agents in DERA then converse with one another, visible in \methodname~dialog. The \Decider adds accepted suggestions to a scratchpad, which collects the final changes to make to the care plan. The final care plan is generated by the \Decider using this scratchpad. Note the points in \textbf{bold} that were added to the final care plan.}\label{fig:cp_ex}
\end{figure*}
\section{Care Plan Generation}
\label{sec:careplan}
We also analyze the performance of \methodname~on the task of generating a care management plan. This care plan contains suggestions that are meant to be \textit{physician-facing} - that is, we generate suggestions that a physician would be required to approve of and then communicate to a patient. Our care plans contain five sections: Medications, Referrals, Tests, Lifestyle, and Supportive Care. 

\begin{figure*}[ht!]
    \centering
    \includegraphics[width=1.0\textwidth]{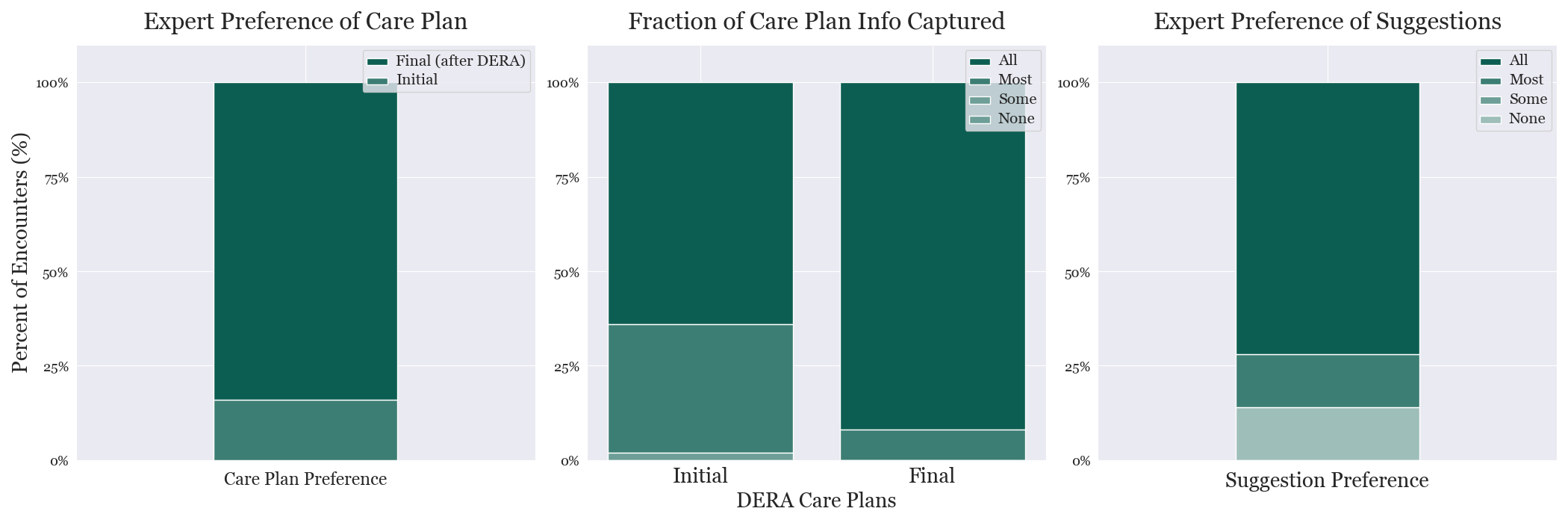}
    \caption{Care plan generation task: Results from physician-expert evaluations. (Left) Physicians choose the final care plan produced by DERA over the initial GPT-4 generated care plan 84\% to 16\%. (Center) Final DERA care plans capture far more of the necessary care management steps than initial GPT-4 generated care plans, with physicians rating "All" relevant steps inferred from the patient-physician chat generated in 92\% of DERA care plans vs. 64\% of initial GPT-4 care plans. (Right) For care plan correction suggestions in the scratchpad, physicians rate agreement with "All" suggestions in 72\% of encounters, Most" in 14\%, "Some" in 0\%, and "None" in 14\%.}
    \label{fig:results_careplan}
\end{figure*}

\paragraph{\methodname~setup} As in the medical conversation summarization task, the goal of DERA is to improve the quality of the generated care plan by suggesting more appropriate home care for the patient, recommending additional lab tests, or otherwise better aligning the generated summary. The DERA setup is the same as the  medical conversation summarization task with care plan-specific prompts. The \Decider starts with an initial care plan. The \Researcher is prompted (Prompt \ref{prompt:cp_researcher}) to converse with the \Decider (Prompt \ref{prompt:cp_decider}). Finally, the \Decider generates the final care plan (Prompt \ref{prompt:cp_final}).  by combining the initial care plan with the content of the `scratchpad' accumulated during the conversation. 

We run DERA on the care plan generation task using GPT-4 with the settings mentioned in Table \ref{tab:summ_experimental_details}.

\paragraph{Dataset} We used the same set of 50 medical encounters we used for the human expert evaluation of the medical conversation summarization task. 
 
\paragraph{Human Experts Evaluation} We evaluated the effectiveness of DERA to generate care plans through human evaluation with four licensed physicians.  We explicitly instructed the physician evaluators that the generated plan is defined as "meant to be provider-facing, meaning that not all suggested interventions will necessarily be recommended to the patient or followed by the patient."  The physicians who evaluated the quality of these care plans were not those who provided care to the patients in the original encounter.

The experts were provided with the
encounter and the two careplans -- the baseline GPT-4 generated summary and the DERA generated summary starting from GPT-4 generated summary. They were asked to answer the following three questions: 

\begin{enumerate}
    \item Which careplan do you prefer to use for the given patient and encounter?  (Options: Initial, Final).
    \item What fraction of the necessary care plan management steps are captured? (Options: All, Most, Some, None)
    \item What percentage of the suggestions added to the DERA scratchpad do you agree with? (Options: All, Most, Some, None)
\end{enumerate}

Figure \ref{fig:results_careplan} shows the results. In a head-to-head comparison, the physicians prefer the final care plan produced by \methodname~84\% of the time.  Furthermore, when asked to give what fraction of care plan corrections were useful, they fully agreed with 72\% of suggestions. They agree with none of the suggestions only 14\% of the time.  Finally, they rated 92\% of care plans as complete, compared to 64\% of initial care plans.  In summation, the application of \methodname~to care plan generation increased the resulting quality substantially.

In addition to these questions, we also asked the physician-experts the following: \textit{If this care plan were acted upon by the patient, does this care plan contain information that could potentially be harmful to the patient given their presentation? } (Options: Yes, No). The amount of careplan containing "harmful" information drops from 2\% in the initial careplan to 0\% in the final DERA summary.  We caution against drawing generalizations from these harmfulness numbers. Our evaluations are both limited in number and drawn from a patient population specific to the telehealth platform; thus cannot predict the generalizability of these findings in other settings.

\paragraph{Qualitative Examples} We show a qualitative example of the care plan generation task with DERA in Figure \ref{fig:cp_ex}. The initial care plan generated by the \Decider was originally rated as containing "Most" necessary care management steps by our physician-expert evaluator, suggesting there were still some improvements possible. In the DERA dialog, the \Researcher highlights potential drug interactions with the patient's current medications and the recommendation to educate the patient on safe sexual practices. These corrections were accepted by the \Decider, as evidenced by the notes written to the scratchpad.  In turn, the corrections were manifested in the final care plan, with the three changes \textbf{bolded}. This final care plan was rated as containing "All" necessary care management steps by our physician-expert evaluator.

\section{Open-Ended Medical Question Answering}
\label{sec:qa}

\paragraph{Overview}

We also investigate the use of \methodname~for short-form medical reasoning.  A commonly used dataset for this task is MedQA \cite{jin2021disease} which consists of USMLE-style practice multiple-choice questions. Previous approaches for this dataset have included using RoBERTa \cite{liu2019roberta}, refining chain-of-thought using GPT-3 \cite{lievin2022can},  and fine-tuning PaLM \cite{chowdhery2022palm, singhal2022large}. While most previously-reported results achieved passing results, recent GPT-4 is shown to work at a near-expert level \cite{nori_capabilities_2023}. 

 In all previous work, the primary focus was on the multiple-choice question format which has limited applicability in the real world.  If these models are to support doctors in decision-making, these models need to operate without any options provided. To mimic this setting, we extend the MedQA dataset to be open-ended to evaluate the model in a more realistic and harder setting.  In an open-ended form, the model must generate the correct answer free-form and not choose from a given bank of options. We also evaluate a set of continuing education questions from the New England Journal of Medicine (NEJM), again in an open-ended setting. 

 A method that can perform at a high level on this task requires several attributes.  First, it must be able to recall a large set of knowledge across multiple domains of medicine.  Second, it must be able to reason over long questions, which will likely include both irrelevant and crucial facts needed to arrive at the solution.

\paragraph{Datasets}

We evaluate our approach using two Medical Question answering datasets - MedQA US dataset \cite{jin2021disease} and New England Journal of Medicine Test Questions (NEJM). Both datasets consist of questions taken from practice or real medical exams (United States Medical Licensing for MedQA, and continuing education questions for NEJM).  For both datasets, the questions are originally written in multiple-choice format (\eg \textit{Which of the following is the best diagnosis?}). Our goal is to test \methodname~'s performance on open-ended question answering, where the task will be to generate the answer free-form.

Therefore, we use \LLM~to alter the questions to be open-ended.  In most cases, this requires a simple rephrasing of the final sentence.  For example, the previous question could be re-written as \textit{What is the best diagnosis?}.  In these cases, we restrict \LLM~to rewrite only the final sentence of the question, so as to guard against hallucinations.  When a more complex rewrite is required, we prompt \LLM~to rewrite the entire question and find that it only changes the relevant sentence.  Some questions could already be answered open-ended and required no rewriting.  Although we performed quality checks, as the entire process is automated, there may be some errors.  The prompts for rewriting the final sentence \ref{prompt:rewrite_last} and the full question \ref{prompt:rewrite_full} are included in the Appendix. We also release the full MedQA open-ended dataset at \url{https://github.com/curai/curai-research/tree/main/DERA}.  We cannot release the NEJM dataset due to licensing issues.  

For MedQA, we sample a portion of the training set (1178 questions) as a development set and maintain the integrity of the test set (1273 questions) as formulated by the authors.  For NEJM, we split the datasets by area, reserving 7 areas \footnote{Reproductive, Gastrointestinal, Neurologic/Psychogenic, Special Sensory, Endocrine, Musculoskeletal, and Maternity Care} as a development set (consisting of 639 questions), with the remainder serving as a test set (1112 questions). We do not exclude questions containing images.
\begin{figure*}[tb]
\small
\begin{tcolorbox}

\textit{\textbf{Question}} [ID 4]:A 20-year-old woman presents with menorrhagia for the past several years. She says that her menses “have always been heavy”, and she has experienced easy bruising for as long as she can remember. Family history is significant for her mother, who had similar problems with bruising easily. The patient's vital signs include: heart rate 98/min, respiratory rate 14/min, temperature 36.1°C (96.9°F), and blood pressure 110/87 mm Hg. Physical examination is unremarkable. Laboratory tests show the following: platelet count 200,000/mm3, PT 12 seconds, and PTT 43 seconds. What is the most likely cause of this patient's symptoms? 

\textit{\textbf{\textcolor{goldanswer}{Gold Answer: Von Willebrand Disease}, \textcolor{oneshotanswer}{1-shot : Inherited Bleeding Disorder}}}
\vspace{-1em}
\begin{chat}
\who{\Decider}
The patient's symptoms of menorrhagia and easy bruising, along with a family history of similar problems, suggest an inherited bleeding disorder. However, Von Willebrand disease, which is the most common inherited bleeding disorder, cannot be ruled out. Both options are possible causes of the patient's symptoms, but the presence of a family history of bleeding problems slightly increases the likelihood of an inherited bleeding disorder.
\who{\Researcher} Remember to consider the fact that the patient has experienced menorrhagia for several years and easy bruising for as long as she can remember, along with a family history of similar problems
\who{\Decider}  After considering the patient's symptoms of menorrhagia and easy bruising, along with a family history of similar problems, I believe that Von Willebrand disease is the most likely cause of this patient's symptoms.
\who{\Researcher} Take note of the laboratory test results, specifically the platelet count, PT, and PTT values, as these can help determine the cause of the patient's symptoms.
\who{\Decider} After taking into consideration the laboratory test results, specifically the platelet count, PT, and PTT values, I still believe that Von Willebrand disease is the most likely cause of this patient's symptoms. The patient's symptoms of menorrhagia and easy bruising, along with a family history of similar problems, further support this diagnosis.
\who{\Researcher} Pay attention to the patient's vital signs, specifically the heart rate, respiratory rate, temperature, and blood pressure, as these can also provide clues to the cause of her symptoms.
\who{\Decider}  After taking into consideration the patient's vital signs, specifically the heart rate, respiratory rate, temperature, and blood pressure, I still believe that Von Willebrand disease is the most likely cause of this patient's symptoms. The patient's symptoms of menorrhagia and easy bruising, along with a family history of similar problems and laboratory test results further support this diagnosis.

\textit{\textbf{\textcolor{deraanswer}{\methodname~ answer: Von Willebrand Disease}}}
\end{chat}

\end{tcolorbox}
\caption{We provide several examples of chats between the \Decider and \Researcher agents taken from our MedQA development set (more shown in Appendix Section \ref{app:qa_examples}). We include the \textcolor{goldanswer}{Gold Answer}, \textcolor{oneshotanswer}{1-shot answer}, and \textcolor{deraanswer}{\methodname~answer}. This example is a case where the single-shot answer is not specific enough, and the \methodname~dialog changes the answer to the correct one.}\label{fig:qa_ex}
\end{figure*}

\paragraph{\methodname~setup}
To generate an initial answer for \methodname~ to discuss, we use a single-shot prompt which outputs a short answer (Prompt \ref{prompt:qa_single_shot}). We use a single-shot prompt to ensure a consistent output, which we were unable to achieve with a zero-shot prompt. Earlier work \cite{singhal2022large} has shown that using a self-consistency strategy provides stronger results.  We adopt this approach by running $5$ completions of our single-shot prompt and selecting the answer with the most votes as the \textit{single-shot} answer, and consider this as our baseline\footnote{We do not account for variations in the text, each lexical form is counted separately.}.

Instead of initializing our \decider with a single answer, we provide it with the distribution of votes.  This approach provides \methodname~with the distribution better captures the underlying uncertainty of the model\footnote{This also handles cases where closely related lexical forms receive separate votes, as the \decider output will conclude that the options are similar.}. A prompt (Prompt \ref{prompt:qa_student_exp}) is tasked with writing a reasoning behind the votes, which is used as the initial \decider message. 

Starting with the initial \decider message, both  \decider (Prompt \ref{prompt:qa_student}) and \researcher have access only to the question and their own conversation as they iteratively discuss the problem and attempt to achieve the right answer.
The \researcher can stop the dialogue when they have exhausted all relevant information, otherwise, it is set to end after $n=3$ turns. At each turn, the \decider must state what their current answer is and explain their reasoning, and they may choose to either confirm or change their answer.

We instruct both prompts to act as medical doctors who work at an expert level. To arrive at a final answer, a prompt is given the same information as the original one-shot prompt, with the exception that it is also given the full chat history to use as additional context.  We generate $n=5$ answers and use the most frequently generated answer as our final answer (see Prompt \ref{prompt:qa_final}).

We run \methodname~on open-ended question answering with the parameters noted in Table \ref{tab:qa_experimental_details}. For the multiple-choice setting, we use a very similar configuration.  The primary prompt changes are limited to the fact that \decider is given a set of options and asked to generate the letter (A-D) instead of a short phrase.

\paragraph{Evaluation Metrics} There are two main challenges when evaluating open-ended responses to the questions. Some questions may have multiple correct answers, given that it is no longer constrained to a discrete set of options. More importantly, we need a mechanism to  map the free-form answer to the correct answer text. While we only evaluate against the ground truth correct option, we include metrics that attempt to account for the degree of similarity between the answer text and the correct option.

To identify generated answers that are related to the gold standard answer, we prompt \LLM~ to score the relatedness of the generated and gold answers (Prompt \ref{prompt:qa_sim}). To evaluate a binary prompt that judges if the generated and gold answers are the exactly same, and provides an explanation (Prompt \ref{prompt:qa_exact}). Finally, we evaluate the generated and gold answer similarity using BERTScore (\citet{zhang2019bertscore}, model \texttt{scibert-basevocab-uncased}). While this metric has limitations \cite{hanna-bojar-2021-fine,sun-etal-2022-bertscore}, it is commonly reported for generation tasks.  We do not use a knowledge base such as UMLS\cite{aronson2001effective} based similarity \cite{mcinnes2009umls} as many answers do not map directly to a single entity in a medical knowledge base.

\begin{table}
\begin{tabular}{l|l}
\toprule
Model                      & Accuracy \\ \midrule
PaLM \cite{singhal2022large}                       & 0.676    \\
\citet{nori_capabilities_2023} & 0.814 \\
GPT-4 0-shot             & 0.834    \\
\methodname & 0.840 \\  \bottomrule
\end{tabular}
\caption{Results on the multiple-choice (4-option) version of MedQA.  The GPT-4 0-shot and \methodname~results were generated on a version of \LLM~available in February 2023.}\label{tab:qa_mc}
\end{table}

\begin{table*}
\centering
\begin{tabular}{@{}l|ccc|ccc@{}}
\toprule
     & \multicolumn{3}{c|}{MEDQA} & \multicolumn{3}{c}{NEJM}   \\ 
     	& BERTScore	& GPT-4 Exact & GPT-4 Sim 	& BERTScore & GPT-4 Exact & GPT-4 Sim  \\ \hline
    GPT-4 1-shot & 0.746 & 0.698 & 0.65  & 0.676 & 0.703 & 0.711  \\
      \methodname  & 0.744 & 0.703 & 0.67 & 0.670 & 0.711 & 0.724  \\
\bottomrule
      
\end{tabular}
\caption{Results on the Open-Ended versions of MedQA and NEJM. We evaluate using a variety of metrics, including \LLM~prompts that identify exact matches and similar matches (using a 0-1 scale).  In addition, we calculate the average BERTScore $F_1$ to measure the similarity between the gold and generated answers using a separate model.}
\label{tab:qa}
\end{table*}

\paragraph{Results}
We compare \methodname~to single-shot performance using \LLM, where $n=5$ answers are detected, and the one with the most votes is selected as the answer\footnote{If there are ties, the first completion of the highest-ranking answers is selected.}.  Due to the costs involved with running the experiments, we only report single runs.  We include quantitative results for open-ended question answering in Table \ref{tab:qa}, and for multiple-choice question answering in Table \ref{tab:qa_mc}.

For the multiple-choice results, we find that \LLM~outperforms the best previously published approaches out of the box on MedQA.  This is in line with that reported by \citet{nori_capabilities_2023}, which uses a very similar approach. We suspect that our results are slightly higher due to our use of a self-consistency approach. We do not see significant improvements when applying \methodname~compared to the multiple choice setting.

In the open-ended setting, we see strong performance in both one-shot GPT-4 and \methodname for both NEJM and MedQA.  \citet{lievin2022can} notes that the passing grade for the MedQA test set is 60\%.  For both GPT-4 one-shot and \methodname, we see that GPT-4 Exact Matching is above 60\% and BERTScore and Similarity measures are above 0.6. This marks an impressive ability to generate open-ended answers to questions.  Yet there still exists a gap between open-ended and multiple-choice performance, suggesting opportunities for future work.

Similarly to the multiple choice setting, \methodname~shows small to no improvement over GPT-4, depending on the metric.  The largest gain for \methodname~is in the similarity metric for both MedQA and NEJM, which suggests that \methodname~can lead to answers that are closer to the ground truth.  Examples of the open-ended question-answering chats are included in Figure \ref{fig:qa_ex} and Appendix Section \ref{app:qa_examples}. 

\paragraph{Qualitative Analysis}\label{sec:qa_analysis}

We include the first $10$ examples from the MedQA development set (which we randomly drew from their training set) in Appendix Table \ref{tab:qa_examples}\footnote{These results were generated with an earlier version of GPT-4 available in February 2023.}.  In our analysis of these development examples, we see several patterns.

First, sometimes the agent successfully changes an incorrect answer to the correct answer.  For example, in Question $4$ shown in Figure \ref{fig:qa_ex}, the original answer is \textit{Inherited bleeding disorder}, and \methodname~changes it to the more specific \textit{Von Willebrand Disease}.  In other cases, \methodname~leaves the answer as the same in the original 1-shot generation (\eg Questions 5, 9, 55, 94, 98).  We also note that this does not occur in a majority of cases, as only $542$ of the $1273$ MedQA training examples have the exact same answer between \methodname~and one-shot.

In other cases, such as in Question 54, \methodname~adds additional details to the 1-shot answer (1-shot \textit{Smoking cessation counseling and support} to the \decider's final answer \textit{Assessing for occupational lung disease and providing smoking cessation}.  There are some clear challenges with open-ended question answering that show in both the \methodname~and 1-shot generations.  Specifically, often both give a more general answer than is given in the gold standard answer.  For example, in Question 74, the gold standard answer text is a specific medication (\textit{Deantrolene}), while both \methodname~and 1-shot produce more general answers (\eg \textit{Immediate hospitalization and supportive care}). 

Overall, without the inclusion of a specific set of options, it is difficult for \LLM~to generate an answer at a correct level of specificity (\eg a specific treatment instead of a general approach) and a correct length (\eg answering in short answer format instead of long sentences).  In some settings, these attributes may be useful, but it results in a challenging approach to evaluate.  We predict the need for additional work in methods that automatically evaluate the output of large language model-powered tools, given the inherent complexity present.

\section{Discussion and Conclusion}
We introduce a framework for agent-agent dialog called \methodname.  This approach allows agents to focus on specific roles, reducing the need for an LLM to achieve the correct answer in one or two generations. In this setup, we use two types of agents -- \Researcher, tasked with reviewing and selecting information, and \Decider, tasked with integrating that information into the final output. Both discuss the problem in a chat format, with the goal of improving the output of \LLM.  

As found in Sections \ref{sec:summarization} and \ref{sec:careplan}, we find \methodname~improves the quality of the generated text in a variety of metrics. Importantly, this reduces the number of hallucinations and omissions in the resulting text.  This finding is important given the ability of large language models (LLM), in particular GPT-4, to generate text that is fluent but potentially prone to errors.  The ability of \methodname~to identify and correct these hallucinations and omissions is critical when applying these models to real-world scenarios.  A key feature is that the same LLM can be harnessed in both roles. 

We did not find similar improvements in the question-answering task.  As discussed in Section \ref{sec:qa}, \methodname~produced little to no improvement over a GPT-4 baseline.  We suggest this is due to several factors, including the requirement to generate a single, granular answer.  \methodname~often adds information to an answer, which is not helpful for short text generation. These findings, paired with those discussed above, suggest this method is well-suited for longer-generation tasks.

Furthermore, the chat-based format of \methodname~allows for increased interpretability when auditing the results. Even though LLMs such as GPT-4 may achieve high performance in zero-shot or one-shot settings, generating long-form explanations does not provide a granular forum for understanding resulting generations.  Conversely, the chat-based format allows for discussions that are granular and could be verified by an end user for mistakes.  

In the future, this setup could be altered to include human input in the discussion.  Alternatively, different problems may dictate the inclusion of different types of agents.  Overall, we believe that while LLM-based tools are critical in increasing the quality of natural language performance, additional research is required to ensure they are consistent and auditable. 

Finally, we reiterate the need for further research in automated metrics for evaluating LLM output.  Human-led qualitative evaluations can provide important insights, but it remains a challenge to measure improvement given the limited tools currently available.

\section{Limitations}
The experiments in this paper were performed using OpenAI's API, mostly using GPT-4 models.  While these models generate text at a higher quality than other previous models, there are still limitations.  First, we do not have access to what the model has and has not been trained on.  Specifically, we do not know if openly-released datasets, such as MedQA, were included in the training data. Second, we report results using the latest version of GPT-4 available at the time.  As OpenAI does not persist models, this may make reproducing results challenging.  

While we include a variety of quantitative evaluations, the task of automatically evaluating generated text needs further research. Previous methods, such as BERTScore, use models that are less powerful than GPT-4, yet using GPT-4 to evaluate itself is also potentially problematic.  Similarly, evaluations of the ability of \methodname~to reduce the presence of harmful text in generations are promising, but given the limited amount of harmful content present to begin with, we caution against a broad interpretation of that result.

Another limitation is inherent to clinical text datasets. As stated in Section \ref{sec:summarization}, our evaluations and style of summary and care plan are limited to a patient population specific to a single telehealth service, and may not be applicable to uses for the general population. Several of our evaluation datasets cannot be openly released for data privacy or licensing reasons, yet this is likely true for a broad amount of medically-focused research. Additionally, some other datasets that are openly available cannot be directly used with API-based models \cite{Agrawal2022LargeLM}, further limiting options. We also acknowledge that while MedQA does probe medical knowledge, it likely does so in a different form than is likely to be applied in a regular clinical setting.


\bibliography{main,anthology}
\bibliographystyle{acl_natbib}

 \appendix

\clearpage
\section{Appendix}

\subsection{Question Answering Examples}
\label{app:qa_examples}

We provide several examples of chats between the \Decider and \Researcher agents taken from our MedQA development set. We include the \textcolor{goldanswer}{Gold Answer}, \textcolor{oneshotanswer}{1-shot answer}, and \textcolor{deraanswer}{\methodname~answer}.  In the following example, the 1-shot prompt arrives at the correct answer, but \methodname~ expands the answer to include additional information.
\noindent\rule{0.48\textwidth}{1pt}
\textit{\textbf{Question}} [ID 574]:Your colleague has been reading the literature on beta-carotene supplementation and the risk of heart disease. She thinks they may share a clinically relevant association and would like to submit an editorial to a top journal. Upon final literature review, she discovers a newly published study that refutes any association between beta-carotene and heart disease. Your colleague is upset; you suggest that she, instead, mathematically pool the results from all of the studies on this topic and publish the findings. What type of study design are you recommending to your colleague?
\textit{\textbf{\textcolor{goldanswer}{Gold Answer: Meta-analysis}, \textcolor{oneshotanswer}{1-shot : Meta-analysis}}}
\vspace{-1em}
\begin{chat}
\who{\Decider}
Both options are viable for pooling results from multiple studies, but a meta-analysis (80\%) is more likely to be recommended as it allows for a quantitative synthesis of the data. A systematic review and meta-analysis (20\%) would also be a good option, as it includes a comprehensive literature search and qualitative analysis, but it may be more time-consuming. Ultimately, the choice depends on the specific goals and resources of the researcher.
\who{\Researcher} Consider the fact that your colleague wants to 'mathematically pool the results from all of the studies on this topic' and think about which study design would best allow for this type of analysis.
\who{\Decider} Based on the fact that my colleague wants to mathematically pool the results from all of the studies, I believe a meta-analysis would be the best study design as it allows for a quantitative synthesis of the data.
\who{\Researcher} Think about the goal of your colleague, which is to 'submit an editorial to a top journal' and consider which study design would be most appropriate for achieving this goal.
\who{\Decider} After considering the goal of my colleague to submit an editorial to a top journal, I believe a systematic review and meta-analysis would be the most appropriate study design. This design includes both a comprehensive literature search and qualitative analysis, in addition to the quantitative synthesis of data, which would make it more suitable for publication in a top journal.
\who{\Researcher} Consider the fact that your colleague discovered a 'newly published study that refutes any association between beta-carotene and heart disease' and think about how this might impact the choice of study design.
\who{\Decider} After considering the fact that my colleague discovered a newly published study that refutes any association between beta-carotene and heart disease, I still believe a systematic review and meta-analysis would be the most appropriate study design. This design would allow for the inclusion of this new study in the comprehensive literature search and analysis, providing a more complete picture of the current state of research on this topic.
\textit{\textbf{\textcolor{deraanswer}{\methodname~ answer: Systematic review and meta-analysis}}}
\end{chat}
In the following example, the 1-shot prompt arrives at the wrong answer, and while \methodname~ changes the answer, it is also not correct.
\noindent\rule{0.48\textwidth}{1pt}
\textit{\textbf{Question}} [ID 2599]:A 15-year-old girl is brought to her pediatrician's office complaining of frequent diarrhea, fatigue, and inability to gain weight. Her vital signs are within normal limits, and her BMI is 17. She describes her stools as pale, malodorous, and bulky. She often has abdominal bloating. Her symptoms are most prominent after breakfast when she typically consumes cereal. After several weeks of careful evaluation and symptomatic treatment, the pediatrician recommends an esophagogastroduodenoscopy. A diagnostic biopsy shows blunting of intestinal villi and flat mucosa with multiple intraepithelial lymphocytes. What is the patient likely deficient in?
\textit{\textbf{\textcolor{goldanswer}{Gold Answer: IgA}, \textcolor{oneshotanswer}{1-shot : Gluten tolerance}}}
\vspace{-1em}
\begin{chat}
\who{\Decider}
Based on the patient's symptoms and the results of the esophagogastroduodenoscopy, it is likely that the patient has a gluten intolerance or Celiac disease. However, without further testing and confirmation, it is difficult to determine the exact diagnosis with certainty. Therefore, the options of Gluten intolerance/Celiac Disease, Gluten intolerance or celiac disease, and Gluten intolerance or Celiac disease are all rated at 20\%. The option of Gluten tolerance is rated at 40\% as it is possible that the patient's symptoms are caused by another condition.
\who{\Researcher} Consider the patient's symptoms of frequent diarrhea, fatigue, and inability to gain weight, along with her description of pale, malodorous, and bulky stools.
\who{\Decider} After considering the patient's symptoms of frequent diarrhea, fatigue, and inability to gain weight, along with her description of pale, malodorous, and bulky stools, I believe that Celiac Disease is the most likely diagnosis. The results of the esophagogastroduodenoscopy showing blunting of intestinal villi and flat mucosa with multiple intraepithelial lymphocytes also support this conclusion.
\who{\Researcher} Think about the patient's symptoms being most prominent after breakfast when she typically consumes cereal, and how this could be related to her condition.
\who{\Decider} After considering the patient's symptoms being most prominent after breakfast when she typically consumes cereal, I believe this further supports my conclusion that Celiac Disease is the most likely diagnosis. The consumption of cereal, which often contains gluten, could be triggering her symptoms.
\who{\Researcher} Consider the diagnostic biopsy results which show blunting of intestinal villi and flat mucosa with multiple intraepithelial lymphocytes, and how this could be related to the patient's condition.
\who{\Decider} After considering the patient's symptoms, the timing of her symptoms being most prominent after consuming cereal, and the diagnostic biopsy results showing blunting of intestinal villi and flat mucosa with multiple intraepithelial lymphocytes, I believe Celiac Disease is the most likely diagnosis.
\textit{\textbf{\textcolor{deraanswer}{\methodname~ answer: Celiac Disease}}}
\end{chat}

\clearpage

\onecolumn
\begin{longtable}{rp{0.4\textwidth}p{0.12\textwidth}p{0.12\textwidth}p{0.12\textwidth}}
\caption{Examples from the MedQA Training set (used as a development set in our experiments).  We include the id (or line number), the rewritten open-ended question, the correct original gold answer, the student predicted answer, and the 1-shot answer.}\label{tab:qa_examples}\\    

\toprule
 id &  question & gold text & \methodname~answer & 1-shot answer \\
\midrule
\endfirsthead

\toprule
 id &                                                                                                                                                                                                                                                                                                                                                                                                                                                                                                                                                                                                                                                                                                                                                                                                                                                                                                                                                                                                                                 question &                      answer text &                                                                                                                 \methodname~answer &                                                 1-shot answer \\
\midrule
\endhead
\midrule
\multicolumn{5}{r}{{Continued on next page}} \\
\midrule
\endfoot

\bottomrule
\endlastfoot
  4 &                                                                                                                                                                                                                                                                                                                                                                                        A 20-year-old woman presents with menorrhagia for the past several years. She says that her menses “have always been heavy”, and she has experienced easy bruising for as long as she can remember. Family history is significant for her mother, who had similar problems with bruising easily. The patient's vital signs include: heart rate 98/min, respiratory rate 14/min, temperature 36.1°C (96.9°F), and blood pressure 110/87 mm Hg. Physical examination is unremarkable. Laboratory tests show the following: platelet count 200,000/mm3, PT 12 seconds, and PTT 43 seconds. What is the most likely cause of this patient's symptoms? &           Von Willebrand disease &                                                                                                       Von Willebrand disease   &                                            Inherited bleeding disorder \\
  5 & A 40-year-old zookeeper presents to the emergency department complaining of severe abdominal pain that radiates to her back, and nausea. The pain started 2 days ago and slowly increased until she could not tolerate it any longer. Past medical history is significant for hypertension and hypothyroidism. Additionally, she reports that she was recently stung by one of the zoo’s smaller scorpions, but did not seek medical treatment. She takes aspirin, levothyroxine, oral contraceptive pills, and a multivitamin daily. Family history is noncontributory. Today, her blood pressure is 108/58 mm Hg, heart rate is 99/min, respiratory rate is 21/min, and temperature is 37.0°C (98.6°F). On physical exam, she is a well-developed, obese female that looks unwell. Her heart has a regular rate and rhythm. Radial pulses are weak but symmetric. Her lungs are clear to auscultation bilaterally. Her lateral left ankle is swollen, erythematous, and painful to palpate. An abdominal CT is consistent with acut... &                   Scorpion sting &                                                                                    Scorpion venom-induced acute pancreatitis   &                              Scorpion venom-induced acute pancreatitis \\
  9 &                                                                                                                                                                                                                                                                                                                                                                                                                                                                                                                                                                                                                                                                                                                                                                  A 35-year-old male presents to his primary care physician with complaints of seasonal allergies. He has been using intranasal vasoconstrictors several times per day for several weeks. What is a likely consequence of the chronic use of topical nasal decongestants? &            Persistent congestion &                                                                             Rhinitis Medicamentosa (rebound nasal congestion)  &                      Rhinitis medicamentosa (rebound nasal congestion) \\
 54 &                                                                                                                                                                                                                                                                                                                                                                                                                                                                                              A 60-year-old man comes to the physician for an examination prior to a scheduled cholecystectomy. He has hypertension treated with hydrochlorothiazide. His mother had chronic granulomatous disease of the lung. He works in a glass manufacturing plant. He has smoked two packs of cigarettes daily for 38 years. His vital signs are within normal limits. Examination shows no abnormalities. Laboratory studies are within the reference range. An x-ray of the chest is shown. What is the most appropriate next step in management? &     Request previous chest x-ray &                                          Assessing for occupational lung disease and providing smoking cessation counseling.   &                              Smoking cessation counseling and support. \\
 55 &                                                                                                                                                                                                                                                                                                                                                                                                                                                                                                                                                                                                                             You are examining a 3-day-old newborn who was delivered vaginally without any complications. The newborn presents with vomiting, hyperventilation, lethargy, and seizures. Blood work demonstrates hyperammonemia, elevated glutamine levels, and decreased blood urea nitrogen. A CT scan demonstrates cerebral edema. Which enzyme defects would result in a clinical presentation similar to this infant? & Carbamoyl phosphate synthetase I &                                                                                             "Urea cycle enzyme deficiencies"   &                                         Urea cycle enzyme deficiencies \\
 64 &                                                                                                                                                                                                                                                                                                                                                                                                                                                                                                                                                                                                                      An 18-year-old man comes to the clinic with his mom for “pins and needles” of both of his arms. He denies any past medical history besides a recent anterior cruciate ligament (ACL) tear that was repaired 1 week ago. The patient reports that the paresthesias are mostly located along the posterior forearms, left more than the right. What would you expect to find on physical examination of this patient? &          Loss of wrist extension & Decreased sensation and possible weakness in both posterior forearms, with the left side being more affected than the right.   &   Decreased sensation and possible weakness in the posterior forearms. \\
 74 &                                                                                                                                                                                                                                                                                                                                                                                                                                                             A 16-year-old girl is brought to the emergency department by her friends who say that she took a whole bottle of her mom’s medication. They do not know which medication it was she ingested. The patient is slipping in and out of consciousness and is unable to offer any history. Her temperature is 39.6°C (103.2°F), the heart rate is 135/min, the blood pressure is 178/98 mm Hg, and the respiratory rate is 16/min. On physical examination, there is significant muscle rigidity without tremor or clonus. What is the best course of treatment for this patient? &                       Dantrolene &                                          Immediate stabilization and supportive care with emergency toxicology consultation.   &                         Immediate hospitalization and supportive care. \\
 77 &                                                                                                                                                                                                                                                                                                                                 A 3-week-old boy is brought to the emergency department by his parents because of a 3-day history of progressive lethargy and difficulty feeding. He was born at term and did not have difficulty feeding previously. His temperature is 39.4°C (103°F), pulse is 220/min, respirations are 45/min, and blood pressure is 50/30 mm Hg. Pulse oximetry on 100\% oxygen shows an oxygen saturation of 97\%. Examination shows dry mucous membranes, delayed capillary refill time, and cool skin with poor turgor. Despite multiple attempts by the nursing staff, they are unable to establish peripheral intravenous access. What is the most appropriate next step in management for this 3-week-old boy? &         Intraosseous cannulation &                                      Establishing intraosseous access for fluid resuscitation and medication administration.   & Intraosseous needle placement for fluid resuscitation and antibiotics. \\
 94 &                                                                                                                                                                                                                                                                                                                                                                                                                                                                                                                                                                                              A 70-year-old man comes to the physician because of a 4-month history of epigastric pain, nausea, and weakness. He has smoked one pack of cigarettes daily for 50 years and drinks one alcoholic beverage daily. He appears emaciated. He is 175 cm (5 ft 9 in) tall and weighs 47 kg (103 lb); BMI is 15 kg/m2. He is diagnosed with gastric cancer. What cytokine is the most likely direct cause of this patient's examination findings? &                             IL-6 &                                                                                          Tumor necrosis factor-alpha (TNF-$\alpha$)   &                                    Tumor necrosis factor-alpha (TNF-$\alpha$) \\
 98 &                                                                                                                                                                                                                                                                                                                                                                                                                                                                                                         Three days after starting a new drug for malaria prophylaxis, a 19-year-old college student comes to the physician because of dark-colored urine and fatigue. He has not had any fever, dysuria, or abdominal pain. He has no history of serious illness. Physical examination shows scleral icterus. Laboratory studies show a hemoglobin of 9.7 g/dL and serum lactate dehydrogenase of 234 U/L. Peripheral blood smear shows poikilocytes with bite-shaped irregularities. What drug has the patient most likely been taking? &                       Primaquine &                                                                                                                    Primaquine  &                                                             Primaquine \\
\end{longtable}

\twocolumn

\begin{table*}[h!]
    \centering
    \begin{tabular}{@{}c|c c c c c c@{}}
    \toprule
        Prompt & temp. & max\_tokens & top\_p & freq. penalty & num. turns \\ \hline
        Summarization - Initial (\ref{prompt:summarization}) & 1 & 512 & 1  & 0 & - \\
        Summarization - Decider (\ref{prompt:summ_decider}) & 1 & 512 & 1 & 0 & 15 \\
        Summarization - Researcher (\ref{prompt:summ_researcher}) & 1 & 512 & 1 & 0 & 15 \\
        Summarization - Corruption (\ref{prompt:summ_corruption}) & 1 & 512 & 1 & 0 & - \\
        Summarization - Final (\ref{prompt:summ_final}) & 1 & 512 & 1 & 0 & - \\
        GPT-F1 Metric - Concept Extractor (\ref{prompt:metric_extractor}) & 0 & 200 & 1 & 0 & - \\
        GPT-F1 Metric - Concept Verifier (\ref{prompt:metric_verifier}) & 0 & 200 & 1 & 0 & - \\
        Care Plan - Initial (\ref{prompt:careplan}) & 1 & 512 & 1 & 0 & - \\
        Care Plan - Decider (\ref{prompt:cp_decider}) & 1 & 512 & 1 & 0 & 15 \\
        Care Plan - Researcher (\ref{prompt:cp_researcher}) & 1 & 512 & 1 & 0 & 15 \\
        Care Plan - Final (\ref{prompt:cp_final}) & 1 & 512 & 1 & 0 & - \\
        \bottomrule
    \end{tabular}
    
    \caption{Experimental details for Summarization and Care Plan Experiments.  For each prompt, we include the Temperature, Maximum Number of Tokens for Generation, Top-P, the Frequency Penalty, and the Number of Turns for the prompt(s) (where applicable).  In all cases, the final parameters of the model were decided by qualitative evaluation of the output.}
    \label{tab:summ_experimental_details}
\end{table*}

\begin{table*}[h!]
    \centering
    \begin{tabular}{@{}c|c c c c c c@{}}
    \toprule
        Prompt & temp. & max\_tokens & top\_p & num comp. & freq. penalty & num. turns \\ \hline
        1-shot (\ref{prompt:qa_single_shot}) & 0.7 & 50 & 1 & 5 & 0 & 1 \\
        Decider Initial Msg. (\ref{prompt:qa_student_exp}) & 0 & 400 & 1 & 1 & 0 & 1\\
        Res.-Dec. Msg. (\ref{prompt:qa_student} \& \ref{prompt:qa_teacher}) & 0.3 & 250 & 1 & 1 & 0.5 & 3 \\
        Final Decider Answer (\ref{prompt:qa_final}) & 0 & 100 & 1 & 5 & 0 & 1 \\
        Pred - Gold Ans. Sim. (\ref{prompt:qa_sim}) & 0 & 100 & 1 & 1 & 0 & 1 \\

        \bottomrule
    \end{tabular}
    \caption{Experimental details for Question Answering.  For each prompt, we include the Temperature, Maximum Number of Tokens for Generation, Top-P, the Number of Completions (\eg the number of generations we prompted from \LLM), the Frequency Penalty, and the Number of Turns for the prompt(s).  In all cases, the final parameters of the model were decided by qualitative evaluation of the output.  For GPT-3 CoT, the two turns refers to 1) the generation of the chain of thought, and 2) the generation of the answer.}
    \label{tab:qa_experimental_details}
\end{table*}

\begin{figure*}[p]
\begin{tcolorbox}[boxsep=1pt,left=2pt,right=2pt,top=0pt,bottom=0pt]
\textbf{Doctor-Patient Chat}
\small
\vspace{-2em}
\begin{chat}
\who{Patient} I have a sinus infection and need something to knock it out. 
\who{Doctor} Hi NAME\, thank you for starting a visit.  I am so sorry to hear about your sinus infection. How long have you had the symptoms for? 
\who{Patient} Since Sunday 
\who{Doctor} Ah I see. Which symptoms do you have at present? 
\who{Patient} My face is swollen my cheeks hurt my eyelids are swollen and I am running a slight fever and I can feel something draining down the back of my throat\ldots
\end{chat}
\vspace{-2em}
\end{tcolorbox}
\begin{tcolorbox}[boxsep=1pt,left=2pt,right=2pt,top=0pt,bottom=0pt]
\textbf{Initial Summary, Corruptions in \textcolor{red}{Red}}

\small
\textit{Demographics and Social Determinants of Health}: The patient's specific demographics are not mentioned in the encounter. The patient experiences sinus infections multiple times a year\ldots

\textit{Patient Intent}: The patient seeks care for a sinus infection, which they have identified themselves. They are \ldots

\textit{Pertinent Positives}: The patient has been experiencing sinus infection symptoms since Sunday. Their face is swollen, cheeks hurt, eyelids are swollen, and they have a slight fever at night. The patient can also feel something draining down the back of their throat. They have a history of 3-6 sinus infections a year, with the \textbf{last one occurring a couple of months ago} $\rightarrow$  \textcolor{red}{\textbf{last two occurring simultaneously a couple of months ago}}.

\textit{Pertinent Unknowns}: \textbf{There are no pertinent unknowns mentioned in this medical encounter.} $\rightarrow$ \textcolor{red}{\textbf{The patient is unsure whether their current condition is a sinus infection or a viral infection.}}.

\textit{Pertinent Negatives}: \textbf{The patient has not been vaccinated against COVID-19} $\rightarrow$ \textcolor{red}{\textbf{The patient has received the COVID-19 vaccine.}}. Their temperature is currently normal but spikes at night. 

\textit{Medical History}: \textbf{The patient has a history of seasonal allergies and sinus infections} $\rightarrow$ \textcolor{red}{\textbf{ The patient has had chickenpox, seasonal allergies, and sinus infections}}. They typically use an Albuterol inhaler for asthma due to their allergies causing wheezing in the spring and fall, however they currently do not have one. They have had COVID-19 before and were previously prescribed \textbf{amoxicillin} $\rightarrow$ \textcolor{red}{\textbf{penicillin}} for sinus infections. \textbf{The patient currently uses Tylenol severe sinus medication and ibuprofen to alleviate symptoms} $\rightarrow$ \textcolor{red}{\textbf{The
patient currently uses ibuprofen and a nasal spray to alleviate symptoms.}}.
\end{tcolorbox}

\begin{tcolorbox}[boxsep=1pt,left=2pt,right=2pt,top=0pt,bottom=0pt]
\textbf{\methodname~dialog}
\small
\vspace{-2em}
\begin{chat}
\who{\Researcher} Let's go through the summary section by section. First, the Pertinent Positives section - there is no mention of the patient taking Tylenol severe sinus medication, which was mentioned in the dialogue. Also, the last sentence about the last two sinus infections occurring simultaneously might be misleading as it's not explicitly mentioned in the dialogue. Could you please add the Tylenol information and consider rephrasing the last part?
\who{\Decider} Thank you for pointing that out. I'll add the information about Tylenol severe sinus medication to the summary. I agree that the last part about the last two sinus infections occurring simultaneously can be misleading; I will rephrase it. Do you have any other suggestions or corrections to make in the summary?
\who{\Researcher} Great, let's move on to the Pertinent Unknowns section. The patient actually mentioned that they know it's a sinus infection, so we shouldn't include that they are unsure about their condition. Could you please remove this part from the Pertinent Unknowns section?
\who{\Decider} You're right, the patient did mention that they know it's a sinus infection. I apologize for the confusion. I'll remove that part from the Pertinent Unknowns section. \ldots
\end{chat}
\vspace{-2em}
\end{tcolorbox}

\begin{tcolorbox}[boxsep=1pt,left=2pt,right=2pt,top=0pt,bottom=0pt]
\textbf{Scratchpad}
\small

Add Tylenol severe sinus medication to the summary, and rephrase the sentence about the last two sinus infections. 

Remove the patient being unsure about their condition from Pertinent Unknowns section. 

Correct the COVID-19 vaccine information in the Pertinent Negatives section. 

Update the summary to mention the patient using the Albuterol inhaler for wheezing and change medication name to amoxicillin.
\end{tcolorbox}
\begin{tcolorbox}[boxsep=1pt,left=2pt,right=2pt,top=0pt,bottom=0pt]
    \textbf{Final Summary}

\small
\textit{Demographics and Social Determinants of Health}: The patient's specific demographics are not mentioned in the encounter. The patient experiences sinus infections multiple times a year\ldots

\textit{Patient Intent}: The patient seeks care for a sinus infection, which they have identified themselves. They are \ldots

\textit{Pertinent Positives}: The patient has been experiencing sinus infection symptoms since Sunday. Their face is swollen, cheeks hurt, eyelids are swollen, and they have a slight fever at night. The patient can also feel something draining down the back of their throat. They have a history of 3-6 sinus infections a year, with the last one occurring a couple of months ago and being treated with amoxicillin. 

\textit{Pertinent Unknowns}: There are no pertinent unknowns mentioned in this medical encounter. 

\textit{Pertinent Negatives}: The patient has not been vaccinated against COVID-19. Their temperature is currently normal but spikes at night. 

\textit{Medical History}: The patient has a history of seasonal allergies and sinus infections. They typically use an Albuterol inhaler for wheezing due to their allergies causing wheezing in the spring and fall, however they currently do not have one. They have had COVID-19 before. The patient currently uses Tylenol severe sinus medication and ibuprofen to alleviate symptoms.
\end{tcolorbox}

\caption{Medical conversation summarization -  Quantitative evaluation: The above is a condensed example taken from our summarization dataset.  Starting with the initial chat, we generate an initial summary, which is then corrupted (see \textcolor{red}{red text}) using Prompt \ref{prompt:summ_corruption}.  The \methodname~dialog generates a scratchpad of edits, which results in the final corrected summary. The quantitative evaluation compares the initial summary with the final summary .}\label{fig:summ_ex}
\end{figure*}

\begin{lstlisting}[language=promptlanguage,float=*,caption=Prompt for generating initial summary.,label={prompt:summarization}]
----
Below is a medical encounter between an {age} 
 and {sex} patient and a doctor done over chat. 
Chief Complaint: "{cc}".
----
Medical Encounter
----
{chat}
----
Summary Instructions
----
Provide a summary of the medical encounter between the doctor and the {age_and_sex} patient in 6 sections (Demographics and Social Determinants of Health, Patient Intent, Pertinent Positives, Pertinent Unknowns, Pertinent Negatives, Medical History). The definitions of each section are listed below. Write a paragraph under each section, not bullet points.

Demographics and Social Determinants of Health: 
// Definition of section

Patient Intent: 
// Definition of section

Pertinent Positives: 
// Definition of section

Pertinent Unknowns:
// Definition of section

Pertinent Negatives: 
// Definition of section

Medical History: 
// Definition of section

----
Summary of Medical Encounter
----
\end{lstlisting}

\begin{lstlisting}[language=promptlanguage,float=*,caption=Prompt for generating corruptions based off of the initial summary.,label={prompt:summ_corruption}]
---
Below is a medical encounter between a {age_and_sex} patient and a doctor done over chat. 
Chief complaint: "{cc}".
----
Medical Encounter
----
{chat}
----
Below is a summary of the conversation that was written using the following instructions:

// Definition of medical summary (same as in initial summarization prompt)
----
Summary of Medical Encounter
----
{summary}
----
Using the above dialogue and provided summary, corrupt the summary slightly. This could include moving a positive symptom to be a negative symptom, making up medical history mentioned, etc.

Corruptions should only occur on the Pertinent Positives, Pertinent Unknowns, Pertinent Negative, or Medical History section.

The lower the desired corruption level, the fewer the changes made. Note that a 0 would be not changing the summary at all, and a 10 would be completely corrupting the summary.

Note that any changes/corruption should make the summary less factual.

Desired Corruption Level: {corruption_level}/10
----
Corrupted Summary of Medical Encounter
----
\end{lstlisting}

\begin{lstlisting}[language=promptlanguage,float=*,caption=Prompt for decider agent\, used in DERA summarization experiments.,label={prompt:summ_decider}]
You (Person A) are a very good summary writer for medical dialogues between physicians and patients.

This is the medical dialogue you summarized for a {age} and {sex} patient:
-Medical Dialogue-
{chat}
-Medical Dialogue-

You are discussing the summary you wrote for this dialogue with another summary writer (Person B) whose job it is to verify your summary for correctness.

Person B will give you points for correction and it will be your job to add the points of correction to a scratchpad if you agree with them.

This is your original version of the summary:
-Your Original Summary-
{summary}
-Your Original Summary-

Here is your current scratchpad of corrections to make to the summary:
-Correction Scratchpad-
{scratchpad}
-Correction Scratchpad-

You are generally very confident about the summary you wrote, however, when presented with compelling arguments by the verifying summary writer, you add to the correction scratchpad. You also suggest any edits of your own in case you notice a mistake.

This is the summary discussion so far:
-Summary Discussion-
{discussion}
-Summary Discussion-

Question: What do you say next? Respond to Person B in the tag [RESPONSE: "<your_response_here>"] and output any corrections to add to the scratchpad in the tag [SCRATCHPAD: "<things_to_add_to_the_scratchpad_here>"]. Make sure to use the "[]" when outputting tags.
Answer:
\end{lstlisting}

\begin{lstlisting}[language=promptlanguage,float=*,caption=Prompt for researcher agent\, used in DERA summarization experiments.,label={prompt:summ_researcher}]
---
You (Person B) are a very good summary editer for medical dialogues between physicians and patients.

This is the medical dialogue you will be referencing for a {age} and {sex} patient:
-Medical Dialogue-
{chat}
-Medical Dialogue-

You are discussing the summary that another summary writer (Person A) wrote for this dialogue one section at a time.

You will be giving Person A points for correction based on any mistakes/discrepancies you see between the dialogue and summary one section at a time. Person A will add the points of correction that they agree on to a scratchpad to later make edits.

However, you will only go through the Pertinent Positives, Pertinent Negatives, Pertinent Unknowns, and Medical History sections.

This is Person A's original version of the summary:
-Person A's Original Summary-
{summary}
-Person A's Original Summary-

Here is Person A's current scratchpad of corrections to make to the summary:
-Correction Scratchpad-
{scratchpad}
-Correction Scratchpad-

Go through each section of the summary one at a time and point out any text that does not have a grounding in the dialogue. It must be possible to directly tie any span of the summary to the dialogue.

Make sure to make accurate, useful suggestions for corrections. 

Person A may not initially agree with you, but if you are confident there is an error do your best to convince Person A of the mistake.

Once you have gone through each section and have confirmed each section with Person A, and you are satisfied with all of the corrections added to the scratchpad and/or all of Person A's reasoning to reject additional corrections, output the tag "[STOP]".

This is the summary discussion with Person A so far:
-Summary Discussion-
{discussion}
-Summary Discussion-

Question: What do you say next? Respond to Person A in the tag [RESPONSE: "<your_response_here>"]. If you are done correcting and are satisfied, output the "[STOP]" tag.
Answer:
\end{lstlisting}

\begin{lstlisting}[language=promptlanguage,float=*,caption=Prompt for final summarization step (incorporating scratchpad of corrections into the original summary)\, used in DERA summarization experiments.,label={prompt:summ_final}]
---
You are a very good summary writer for medical dialogues between physicians and patients.

This is the medical dialogue you summarized for a {age} and {sex} patient:
-Medical Dialogue-
{chat}
-Medical Dialogue-

This is your original version of the summary:
-Original Summary-
{summary}
-Original Summary-

Here is your current scratchpad of corrections to make to the summary:
-Correction Scratchpad-
{scratchpad}
-Correction Scratchpad-

Make all changes mentioned in the scratchpad to the original summary to output the corrected summary.

Output the tag "[STOP]" when finished writing the corrected summary.

-Corrected Summary-
\end{lstlisting}

\begin{lstlisting}[language=promptlanguage,float=*,caption=Prompt for extracting medical concepts from the summary\, used to compute the GPT-F1 metric.,label={prompt:metric_extractor}]
Given the following snippet of a medical dialogue summary, extract the medical concepts (symptoms, diseases, conditions, allergies, lab tests, etc.) present.

The heading of the section from which the summary was extracted will also be provided.

---Example 1---
Pertinent Negatives:  Patient reports no <concept_1>, no <concept_2>, <concept_3>, and <concept_4>. Patient also reports having no trouble with <concept_5>.

Medical Concepts: [<concept_1>, <concept_2>, <concept_3>, <concept_4>, <concept_5>]
---Example 1---

---Example 2---
Pertinent Positives:  Patient ongoing <concept_1> for the past 5 days, <concept_2>, and some <concept_3>. Patient had <concept_4> done in May 2021.

Medical Concepts: [<concept_1>, <concept_2>, <concept_3>, <concept_4>]
---Example 2---

---Example 3---
Pertinent Unknowns:  Patient is unsure about <concept_1> and <concept_2>.

Medical Concepts: [<concept_1>, <concept_2>]
---Example 3---

---Example 4---
Medical History: Patient reports some <concept_1> in the past, and had last <concept_2> on DATE_1.

Medical Concepts: [<concept_1>, <concept_2>]
---Example 4---

Here is the example to extract medical concepts from:

{section_heading}: {section_value}

Medical Concepts:
\end{lstlisting}

\begin{lstlisting}[language=promptlanguage,float=*,caption=Prompt for verifying medical concepts from a summary section\, used to compute the GPT-F1 metric.,label={prompt:metric_verifier}]
Given a snippet (snippet) from a medical dialogue summary and a corresponding list (list_a) of medical concepts extracted from that snippet, evaluate what medical concepts from a separate list (list_b) can be found in either list_a or snippet.

Note that on some occasions a medical concept from list_b may not be found in list_a, but can be appropriate to be present given the snippet. This could include rephrasings of medical concepts that are clinically equivalent (Ex: COVID and COVID-19).

---Example---
snippet: <snippet>
list_a: [<concept_1>, <concept_2>, <concept_3>, <concept_4>, <concept_5>, <concept_7>]
list_b: [<concept_0>, <concept_1>, <concept_3>, <concept_4>, <concept_5>, <concept_6>]

found_b: [<concept_1>, <concept_3>, <concept_4>, <concept_5>]
not_found_b: [<concept_0>, <concept_6>]

---Example---

Here is the snippet, list_a. Evaluate the medical concepts in list_b as above.

snippet: {snippet}
list_a: {list_a}
list_b: {list_b}

found_b:
\end{lstlisting}

\begin{lstlisting}[language=promptlanguage,float=*,caption=Prompt for generating initial care plan,label={prompt:careplan}]
----
Care Plan Instructions
----
You are a primary care physician tasked with writing a care plan, which lists the next steps in care management that the patient and the physician will perform. 
Categorize the next steps into five sections: Medications, Referrals, Tests, Lifestyle and Supportive Care. Definitions and scopes of each section are defined below.

Medications:
// Definition of section
Referrals:
// Definition of section
Tests:
// Definition of section
Lifestyle:
// Definition of section
Supportive Care:
// Definition of section
 
{example}
----
Care Plan Instructions
----
Now that you've seen an example, you will now write a care plan of the same format (five sections: Medications, Referrals, Tests, Lifestyle and Supportive Care). 

The dialogue you will use to write a care plan about is a medical encounter between a {age} and {sex} patient and a doctor done over chat:
----
Dialogue
----
{chat}
----
Care Plan
----
\end{lstlisting}

\begin{lstlisting}[language=promptlanguage,float=*,caption=Prompt for decider agent\, used in DERA care plan experiments.,label={prompt:cp_decider}]
---
You (Person A) are a very good writer of care plans for patients following their discussion with a physician. The full instructions are presented below.
---
Care Plan Writing Instructions
---
// Same instructions as in initial care plan generation prompt. Removed for brevity.
---
Given the instructions, this is the medical dialogue you see for a  {{age}} {{sex}} patient:
---
Medical Dialogue
---
{chat}
---
You are discussing the care plan you wrote for this dialogue with another care plan writer (Person B) whose job it is to verify your care plan for soundness.

Person B will give you points for correction and it will be your job to add the points of correction to a scratchpad if you agree with them.

This is your original version of the care plan:
---
Your Original Care Plan
---
{careplan}
---
Here is your current scratchpad of corrections to make to the care plan:
---
Correction Scratchpad
---
{scratchpad}
---
You are generally very confident about the care plan you wrote, however, when presented with compelling arguments by the verifying care plan writer, you add to the correction scratchpad. You also suggest any edits of your own in case you notice a mistake.

This is the care plan discussion so far:
---
Care Plan Discussion
---
{discussion}
---
Question: What do you say next? Respond to Person B in the tag [RESPONSE: "<your_response_here>"] and output any corrections to add to the scratchpad in the tag [SCRATCHPAD: "<things_to_add_to_the_scratchpad_here>"]. Make sure to use the "[]" when outputting tags. All text should be within the tag brackets.
An example answer would be: [RESPONSE: "I think we should remove ... from the care plan"] [SCRATCHPAD: "Remove ... from the care plan because ..."]
---
Answer:
\end{lstlisting}

\begin{lstlisting}[language=promptlanguage,float=*,caption=Prompt for researcher agent\, used in DERA care plan experiments.,label={prompt:cp_researcher}]
---
You are a primary care physician and very good editor of care plans for patients following their discussion with a physician. The full instructions for writing care plans are presented below.
---
Care Plan Writing Instructions
---
// Same instructions as in initial care plan generation prompt. Removed for brevity.
---
Given the instructions, this is the medical dialogue you see for a {age_and_sex} patient:
---
Medical Dialogue
---
{chat}
---

You are discussing the care plan that another care plan writer (Person A) wrote for this dialogue one section at a time.

You will be giving Person A points for correction based on any reconsiderations you see between the dialogue and care plan one section at a time. Person A will add the points of correction that they agree on to a scratchpad to later make edits.

This is Person A's original version of the care plan:
---
Person A's Original Care Plan
---
{careplan}
---
Here is Person A's current scratchpad of corrections to make to the care plan:
---
Correction Scratchpad
---
{scratchpad}
---
Go through each section of the care plan one section at a time and point out any suggestions that does not have a grounding in the dialogue. All suggestions must be grounded in information from the dialogue.

Remember to make sure the care plan is congruent with the Care Plan Writing Instructions.

Make sure to make accurate, useful suggestions for corrections. 

Person A may not initially agree with you, but if you are confident there is an error do your best to convince Person A of the mistake.

Once you have gone through each section and have confirmed each section with Person A, and you are satisfied with all of the corrections added to the scratchpad and/or all of Person A's reasoning to reject additional corrections, output the tag "[DONE]".

This is the care plan discussion with Person A so far:
---
Care Plan Discussion
---
{discussion}
---
Question: What do you say next? Respond to Person A in the tag [RESPONSE: "<your_response_here>"]. If you are done correcting, are satisfied, and want to end the conversation, output "DONE".
---
Answer:
\end{lstlisting}

\begin{lstlisting}[language=promptlanguage,float=*,caption=Prompt for final care plan generation step (incorporating scratchpad of corrections into the original care plan)\, used in DERA care plan experiments.,label={prompt:cp_final}]
---
You are a very good writer of care plans for patients following their discussion with a physician. The full instructions are presented below.
---
Care Plan Writing Instructions
---
// Same instructions as in initial care plan generation prompt. Removed for brevity.
---
Given the instructions, this is the medical dialogue you see for a {age} and {sex} patient:
---
Medical Dialogue
---
{{chat}}
---
You have been discussing the care plan you wrote for this dialogue with another care plan writer (Person B) whose job it is to verify your care plan for soundness.

You added corrections to a scratchpad after discussing them with Person B, and you will later be tasked with updating the original care plan based off of the correctness suggested in the scratchpad.

This is your original version of the care plan:
---
Your Original Care Plan
---
{careplan}
---
Here is your current scratchpad of corrections to make to the care plan:
---
Correction Scratchpad
---
{scratchpad}
---
Make all changes mentioned in the scratchpad to the original care plan to output the corrected care plan. Make sure all changes are congruent to the Care Plan Writing Instructions.

Output the tag "[STOP]" when finished writing the corrected care plan.
---
Corrected Care Plan
---
\end{lstlisting}

\begin{lstlisting}[language=promptlanguage,float=*,caption=Prompt for rewriting the question in full (temperature at 0 and otherwise uses default parameters),label={prompt:rewrite_full}]
The following question was written as a multiple choice question.  Rewrite it as posing an open-ended question. If it is already an open-ended question and the question requires no rewrite, output "[OPEN]" only.  Do not change any details or facts in the question, and only change the phrasing of the question.

--Example--
Question: A 60-year-old man comes to the physician for an examination prior to a scheduled cholecystectomy. He has hypertension treated with hydrochlorothiazide. His mother had chronic granulomatous disease of the lung. He works in a glass manufacturing plant. He has smoked two packs of cigarettes daily for 38 years. His vital signs are within normal limits. Examination shows no abnormalities. Laboratory studies are within the reference range. An x-ray of the chest is shown. Which of the following is the most appropriate next step in management?

Rewrite: A 60-year-old man comes to the physician for an examination prior to a scheduled cholecystectomy. He has hypertension treated with hydrochlorothiazide. His mother had chronic granulomatous disease of the lung. He works in a glass manufacturing plant. He has smoked two packs of cigarettes daily for 38 years. His vital signs are within normal limits. Examination shows no abnormalities. Laboratory studies are within the reference range. An x-ray of the chest is shown. What is the most appropriate next step in management?
--Example--
Question: Several patients at a local US hospital present with chronic secretory diarrhea. Although there are multiple potential causes of diarrhea present in these patients, which of the following is most likely the common cause of their chronic secretory diarrhea?

Rewrite: Several patients at a local US hospital present with chronic secretory diarrhea.  Although there are multiple potential causes of diarrhea present in these patients, what is most likely the common cause of their chronic secretory diarrhea?

--Example--
Question: A 39-year-old male presents to your office with nodular skin lesions that progress from his right hand to right shoulder. The patient reports that the initial lesion, currently necrotic and ulcerative, developed from an injury he received while weeding his shrubs a couple weeks earlier. The patient denies symptoms of respiratory or meningeal disease. Which of the following most likely characterizes the pattern of this patient's skin lesions:

Rewrite: A 39-year-old male presents to your office with nodular skin lesions that progress from his right hand to right shoulder. The patient reports that the initial lesion, currently necrotic and ulcerative, developed from an injury he received while weeding his shrubs a couple weeks earlier. The patient denies symptoms of respiratory or meningeal disease. How would you characterize the pattern of this patient's skin lesions?
--Example--
Question: A 71-year-old man presents to the clinic with complaints of right wrist pain for 2 days. On examination, redness and swelling were noted on the dorsal aspect of his right wrist. He had  pain with extreme range of motion of the wrist. His history includes 2 hip replacements, 2 previous episodes of gout in both first metatarsophalangeal joints, and hypertension. Two days later, the swelling had increased in the dorsal aspect of his right wrist and hand. Wrist flexion was limited to 80% with severe pain. The pain was present on palpation of the scaphoid bone. Due to the suspicion of fracture, the patient was referred to his general practitioner for radiographs. These findings were consistent with gouty arthritis. What is the most likely cytokine involved in this process?

Rewrite: [OPEN]
---
Question: {{question}}

Rewrite:
\end{lstlisting}

\begin{lstlisting}[language=promptlanguage,float=*,caption=Prompt for rewriting the question by changing the last sentence only (temperature at 0 and otherwise uses default parameters).,label={prompt:rewrite_last}]
The following question was written as a multiple choice quesiton.  For the sentence in the question poses a multiple choice, rewrite it as posing an open-ended question. If the relevant is a compound sentence, re-write the entire sentence.  If it is already an open-ended question and the question requires no rewrite, output "[OPEN]" only.  Do not change any details or facts in the question, and only change the phrasing of the question.

--Example--
Question: A 60-year-old man comes to the physician for an examination prior to a scheduled cholecystectomy. He has hypertension treated with hydrochlorothiazide. His mother had chronic granulomatous disease of the lung. He works in a glass manufacturing plant. He has smoked two packs of cigarettes daily for 38 years. His vital signs are within normal limits. Examination shows no abnormalities. Laboratory studies are within the reference range. An x-ray of the chest is shown. Which of the following is the most appropriate next step in management?

Original: Which of the following is the most appropriate next step in management?

Rewrite: What is the most appropriate next step in management?
--Example--
Question: Several patients at a local US hospital present with chronic secretory diarrhea. Although there are multiple potential causes of diarrhea present in these patients, which of the following is most likely the common cause of their chronic secretory diarrhea?

Original: Although there are multiple potential causes of diarrhea present in these patients, which of the following is most likely the common cause of their chronic secretory diarrhea?

Rewrite: Although there are multiple potential causes of diarrhea present in these patients, what is most likely the common cause of their chronic secretory diarrhea?
--Example--
Question:A 39-year-old male presents to your office with nodular skin lesions that progress from his right hand to right shoulder. The patient reports that the initial lesion, currently necrotic and ulcerative, developed from an injury he received while weeding his shrubs a couple weeks earlier. The patient denies symptoms of respiratory or meningeal disease. Which of the following most likely characterizes the pattern of this patient's skin lesions:

Original: Which of the following most likely characterizes the pattern of this patient's skin lesions:

Rewrite: How would you characterize the pattern of this patient's skin lesions?
--Example--
Question: A 71-year-old man presents to the clinic with complaints of right wrist pain for 2 days. On examination, redness and swelling were noted on the dorsal aspect of his right wrist. He had  pain with extreme range of motion of the wrist. His history includes 2 hip replacements, 2 previous episodes of gout in both first metatarsophalangeal joints, and hypertension. Two days later, the swelling had increased in the dorsal aspect of his right wrist and hand. Wrist flexion was limited to 80% with severe pain. The pain was present on palpation of the scaphoid bone. Due to the suspicion of fracture, the patient was referred to his general practitioner for radiographs. These findings were consistent with gouty arthritis. What is the most likely cytokine involved in this process?

Original: What is the most likely cytokine involved in this process?

Rewrite: [OPEN]
---
Question: {{question}}

Original:
\end{lstlisting}

\begin{lstlisting}[language=promptlanguage,float=*,caption=Prompt for generating the single-shot answer.,label={prompt:qa_single_shot}]
Given the following medical question, respond with the phrase that best answers the question.

--Example--
Question: A mother brings her 3-week-old infant to the pediatrician's office because she is concerned about his feeding habits. He was born without complications and has not had any medical problems up until this time. However, for the past 4 days, he has been fussy, is regurgitating all of his feeds, and his vomit is yellow in color. On physical exam, the child's abdomen is minimally distended but no other abnormalities are appreciated. What embryologic error could account for this presentation?

What phrase best answers the question posed?

Answer: Abnormal migration of ventral pancreatic bud
----
Question: {question}

What phrase best answers the question posed?

Answer:
\end{lstlisting}

\begin{lstlisting}[language=promptlanguage,float=*,caption=Prompt for generating the explanation for the single-shot answer distribution.,label={prompt:qa_student_exp}]
{question}

{options_filtered_str}

You think the relative likelihood of each option is {relative_likelihood}. Write a 3-4 sentence message explaining why you rate the options in that way, without taking a decisive stand.

Message:
\end{lstlisting}

\begin{lstlisting}[language=promptlanguage,float=*,caption=Prompt for question-answering \researcher.,label={prompt:qa_teacher}]
You are an expert medical doctor who is guiding a medical student through thinking about which of several answers is best for a given question.  You cannot give the student the answer.  Your role is to help the student think through the question, specifically by pointing out portions of the question that are important in understanding the problem.
Rules;
- All responses should include a quote from the question.
- Consider what you, as the teacher, have said in the previous conversation, and do not repeat yourself.
- Responses should be at most 4 sentences long.
- Stop only when you, as the teacher, have pointed out all important aspects of the question in the previous discussion.  To stop, respond with 'STOP' at the next turn.
 You cannot;
 - Directly give the answer to the student
 - Include the correct option in your response, or any paraphrasing of the correct answer.
 - Do not narrow down the options in your response.

Question: {question}

The previous discussion between you and the expert advisor is as follows;
{chat_history}
{last_student_message}

Help the student find the correct answer by pointing out specific parts of the questions they need to think through, but do not include the correct phrase in your response. Your response should be no more than 3-4 sentences.  If you have pointed out all challenging aspects of the question in the previous conversation, respond with "STOP" after the student's next turn.

Response:
\end{lstlisting}

\begin{lstlisting}[language=promptlanguage,float=*,caption=Prompt for question-answering \decider.,label={prompt:qa_student}]
You are an expert doctor who is trying to select the answer to a medical question, and is willing to be open-minded about their answer.   The questions are taken from a short-answer medical exam, and your role is to arrive at the correct answer.

You are chatting with an expert medical advisor, who will try to help you think through the problem, but will not directly tell you the answer.  They will help you by pointing out aspects of the question that are important in finding the answer.  Do not assume that the teacher knows the answer; only that they know how to think through the question. You can change your answer at any point, but do not assume that the expert knows the exact answer and is providing leading questions. Think about their guidance as a whole, and do not only respond to their last message

Question: {question}

The previous discussion between you and the expert advisor is as follows;
{chat_history}
{last_teacher_message}

Rethink the question by considering what the teacher pointed out, in light of your original hypothesis.  Remember they do not know the answer, but only how to think through the question. You can change your mind on the correct answer, but remember that unless the question explicitly asks for multiple answers, you can only provide a single answer. Respond with the option you believe most likely to be the right answer ("Answer:<SHORT ANSWER>") and a response to that message ("Response:<MESSAGE>"):

Answer:
\end{lstlisting}

\begin{lstlisting}[language=promptlanguage,float=*,caption=Prompt for question-answering final answer.,label={prompt:qa_final}]
You are an expert doctor who is trying to select the answer to a medical question, and is willing to be open-minded about their answer.   The questions are taken from a short-answer medical exam, and your role is to arrive at the correct answer.

You are chatting with an expert medical advisor, who will try to help you think through the problem, but will not directly tell you the answer.  They will help you by pointing out aspects of the question that are important in finding the answer.  Do not assume that the teacher knows the answer; only that they know how to think through the question. You can change your answer at any point, but do not assume that the expert knows the exact answer and is providing leading questions. Think about their guidance as a whole, and do not only respond to their last message

Question: {question}

The previous discussion between you and the expert advisor is as follows;
{chat_history}
{last_teacher_message}

Rethink the question by considering what the teacher pointed out, in light of your original hypothesis.  Remember they do not know the answer, but only how to think through the question. You can change your mind on the correct answer, but remember that unless the question explicitly asks for multiple answers, you can only provide a single answer. Respond with the option you believe most likely to be the right answer ("Answer:<SHORT ANSWER>") and a response to that message ("Response:<MESSAGE>"):

Answer:
\end{lstlisting}

\begin{lstlisting}[language=promptlanguage,float=*,caption=Prompt similar to that used for similarity score between generated and gold answers.  Note that occasionally this outputs a number outside of 0-1.  Unless these are all 100s\, we set these to 0s.  This commonly occurs with math problems.,label={prompt:qa_sim}]
Assign a dxSimilarityScore to each of the following pairs where the first diagnosis is an "expectedDx" and the second diagnosis is the "providedDiagnosis".

Expected Vs Provided Dx Pairs:
{answer_text} | {predicted_answer_text}
{answer_text} | {zero_shot_option_index}

Output each pair in one line using this format "dx1" "|" "dx2" "|" "dxSimilarityScore"
output:
\end{lstlisting}

\begin{lstlisting}[language=promptlanguage,float=*,caption=Prompt for exact matching between generated and gold answers.,label={prompt:qa_exact}]
Question:{question}

Do the following two answers refer to the same medical concept? Respond with an answer ("Answer:True" or "Answer:False") followed by an explanation ("Explanation:")

{answer_text}
{predicted_answer_text}

Answer:
\end{lstlisting}

\end{document}